\documentclass{article} 
\usepackage{iclr2027_conference, times}

\usepackage[utf8]{inputenc}
\usepackage{hyperref}
\usepackage{url}
\usepackage{threeparttable}

\usepackage{tikz}

\usepackage{mathtools}
\usepackage{algpseudocode}
\usepackage[ruled,,vlined,linesnumbered]{algorithm2e}
\usepackage{booktabs}
\usepackage{amsmath}
\usepackage{amssymb}
\usepackage{amsthm}
\usepackage{multirow, multicol}
\usepackage{lineno}
\usepackage{subcaption} 
\usepackage{stfloats}  
\usepackage{graphicx}
\usepackage{array}
\newlength{\atlastile}
\newlength{\atlasstrip}
\newcommand{\atlasdir}{figs}               
\newcommand{\dslabel}[1]{\rotatebox[origin=c]{90}{\footnotesize\textsc{#1}}}
\newcommand{\pairlabel}[2]{%
  \scriptsize\begin{tabular}[c]{@{}r@{}}#1\\[-2pt]$\downarrow$\\[-2pt]#2\end{tabular}}
\newcommand{\atlasclean}[1]{\includegraphics[width=\atlastile]{\atlasdir/clean/#1.png}}
\newcommand{\atlasadv}[1]{\includegraphics[width=\atlasstrip]{\atlasdir/strips/#1.png}}

\SetAlFnt{\small}
\SetAlCapFnt{\small}

\theoremstyle{definition}

\author{Marsalis Gibson \quad Claire Tomlin \quad Shankar Sastry \\
Department of Electrical Engineering and Computer Sciences \\
University of California, Berkeley \\
\texttt{\{mtgibson, tomlin, shankar\_sastry\}@berkeley.edu}
}

\newcommand{\mgnote}[1]%
    {\textcolor{red}{\textbf{MG: #1}}}

\newcommand{\valuefunc}{\ensuremath{\mathcal{V}}}
\newcommand{\inputsample}{\ensuremath{x}}
\newcommand{\adversarialscore}{\ensuremath{\iota}}
\newcommand{\advfunclandscape}{\ensuremath{f_{\mathrm{adv}}}}
\newcommand{\superlevelset}{\ensuremath{\mathcal{S}}}

\newcommand{\threshold}{\ensuremath{\phi}}
\newcommand{\superlevelprobweight}{w}

\newcommand{\targetsystem}{\ensuremath{\mathcal{T}}}
\newcommand{\adversarialobjective}{\ensuremath{\mathcal{O}}}
\newcommand{\inputspace}{\ensuremath{\mathcal{X}}} 
\newcommand{\targetoutputspace}{\ensuremath{\Omega}}
\newcommand{\genericadversarialset}{\ensuremath{\mathcal{A}}}

\newcommand{\generator}{\ensuremath{G}}
\newcommand{\latentinputsample}{\ensuremath{z}}
\newcommand{\classcondition}{\ensuremath{s}}
\newcommand{\generatoroutput}{\ensuremath{I}}
\newcommand{\numMNISTclasses}{\ensuremath{n_c}}

\newcommand{\unlabeleddataset}{\ensuremath{\bar{\mathcal{D}}}}
\newcommand{\datasetsample}{\ensuremath{\inputsample}}

\newcommand{\calibration}{\ensuremath{\mathcal{C}_{global}}}
\newcommand{\localcalibration}{\ensuremath{\mathcal{C}_{local}}}

\newcommand{\learningalgorithm}{\ensuremath{L}}
\newcommand{\hypothesismodel}{\ensuremath{h}}

\newcommand{\dimension}{\ensuremath{d}}

\newcommand{\compressionset}{\ensuremath{\mathcal{Q}}}

\newcommand{\numtrustregions}{\ensuremath{N}}

\newcommand{\boundaryweight}{\ensuremath{c}} 
\newcommand{\gpm}{\ensuremath{\mathcal{GP}}}   
\newcommand{\localgp}[1]{\gpm_{#1}} 

\newcommand{\mean}[3]{\mu_{#1#2}^{#3}}     
\newcommand{\variance}[3]{\sigma_{#1#2}^{#3}} 

\newcommand{\trustregion}{\ensuremath{\mathcal{TR}}}
\newcommand{\trustregioncenter}{\ensuremath{c}}
\newcommand{\trustregionlength}{\ensuremath{l}}
\newcommand{\trustregionsampledimension}{\ensuremath{\dimension}}
\newcommand{\localsamplingbudget}{\ensuremath{b_L}}

\newcommand{\locallabelingfunc}{\ensuremath{\advfunclandscape}}

\newcommand{\globalmodel}{\ensuremath{\hypothesismodel_{g}}}
\newcommand{\residualmodel}{\ensuremath{\hypothesismodel_{r}}}
\newcommand{\calibratedmodel}{\tilde{\hypothesismodel}_{\mathrm{g}}}
\newcommand{\localcalibratedmodel}{\tilde{\hypothesismodel}_{\trustregion}}
\newcommand{\genericcalibratedmodel}{\tilde{\hypothesismodel}}

\newcommand{\localuncertainty}{\ensuremath{\variance{i}{}{}}}
\newcommand{\localmean}{\ensuremath{\mean{i}{}{}}}

\newcommand{\globaluncertaintyparam}{\ensuremath{\beta}}
\newcommand{\postprob}{\ensuremath{P}}

\newcommand{\seed}{\ensuremath{\zeta}}
\newcommand{\seedset}{\ensuremath{Z}}
\newcommand{\seedseparation}{\ensuremath{\eta}}

\newcommand{\trustregioncenters}{\ensuremath{\mathcal{C}}}
\newcommand{\boundtrustregioncenters}{\ensuremath{\mathcal{C}_{B}}}
\newcommand{\sattrustregioncenters}{\ensuremath{\mathcal{C}_{sat}}}

\newcommand{\volume}{\ensuremath{V}}

\newcommand{\genericglobalacquisition}{\ensuremath{\mathsf{a}}}
\newcommand{\boundaryglobalacquisition}{\ensuremath{\genericglobalacquisition_{bound}}}
\newcommand{\satglobalacquisition}{\ensuremath{\genericglobalacquisition_{sat}}}

\newcommand{\genericlocalacquisition}{\ensuremath{\bar{\mathsf{a}}}}
\newcommand{\boundarylocalacquisition}{\ensuremath{\genericlocalacquisition_{bound}}}
\newcommand{\satlocalacquisition}{\ensuremath{\genericlocalacquisition_{sat}}}

\newcommand{\localboundarymixingweight}{\rho}

\newcommand{\penaltyfunc}{\ensuremath{g}}
\newcommand{\penaltyfuncinput}{\ensuremath{r}}
\newcommand{\penaltyfuncweight}{\ensuremath{\hat{\alpha}}}

\newcommand{\numboundaryregions}{\ensuremath{\numtrustregions_{bound}}}
\newcommand{\numsatregions}{\ensuremath{\numtrustregions_{sat}}}

\newcommand{\datasetsamples}{\ensuremath{\mathcal{D}_{\mathrm{total}}}}

\newcommand{\certificationsetsize}{\ensuremath{N_{\mathrm{cert}}}}

\newcommand{\globalsamplingbudget}{\ensuremath{b_{G}}}

\newcommand{\expbudget}{\ensuremath{650}}
\newcommand{\numseeds}{\ensuremath{30}}
\newcommand{\toyfunction}{\ensuremath{f}_{toy}}
\newcommand{\toyinputdim}{\ensuremath{\dimension}}
\newcommand{\toyinputidx}{\ensuremath{i}}
\newcommand{\toyinput}{\ensuremath{x}}

\newcommand{\randomoffset}{\ensuremath{p}}
\newcommand{\superlevelratio}{\ensuremath{h_{\text{ratio}}}}
\newcommand{\thresholdsetsize}{\ensuremath{N_{\text{thres}}}}

\DeclareMathOperator{\spread}{Spread}
\DeclareMathOperator{\clump}{Spacing}

\newcommand{\latentSpace}{\ensuremath{\mathcal{Z}}}
\newcommand{\zscore}{\ensuremath{z_{score}}}
\newcommand{\priormean}{\ensuremath{\mu_{latent}}}
\newcommand{\priorsigma}{\ensuremath{\sigma_{latent}}}

\newcommand{\dispersionmetric}{\ensuremath{\spread_{U}}}
\newcommand{\dispersionmetricnorm}{\ensuremath{\spread}}

\newcommand{\logitfunc}{\ensuremath{\widehat{Z}}}

\newcommand{\logitfuncoutput}{\ensuremath{z}}

\newcommand{\ackleyconstanta}{\ensuremath{a}}
\newcommand{\ackleyconstantb}{\ensuremath{b}}
\newcommand{\ackleyconstantc}{\ensuremath{c}}
\newcommand{\rastriginconstanta}{\ensuremath{A}}
\newcommand{\rosenbrockconstanta}{\ensuremath{a}}
\newcommand{\rosenbrockconstantb}{\ensuremath{b}}

\title{ATLAS-AL: Adaptive Trust-Region for Latent
Adversarial Searches via Active Learning}

\iclrfinalcopy 

\begin{document}

\date{}

\maketitle
\begin{abstract}
Security evaluation of learning-based systems requires more than just testing the system against a fixed collection of attacks. It requires adaptive mechanisms that can efficiently discover \textit{sets} of inputs that induce model failure. We introduce ATLAS (Adaptive Trust-Regions for Latent Adversarial Searches), which is a query-based framework that discovers adversarial input sets for black-box learning systems. ATLAS casts attack generation as an active learning level set estimation problem then combines calibrated approximations with a local-global sampling architecture to find regions of the input space that contain adversarial examples. Once discovered, ATLAS is designed to sample points within these adversarial regions to build adversarial sets that accurately represent the state of robustness of the target model. When applied on toy experiments, we find that ATLAS is able to recover more of the adversarial region under a limited query budget than does previous work. When applied to standard and adversarially trained MNIST, CIFAR, and ImageNet model targets, ATLAS produces better representative attacks than other query-based black-box attacks (NES, SignHunter, BayesOpt). ATLAS represents an automated red-teaming framework that can be used for both analyzing the robustness of learning-based systems under development and continuous auditing to see how the robustness of a system changes over time.
\end{abstract}  
\section{Introduction}
\label{Sec: Introduction}
\begin{figure}[t] 
    \centering
    \includegraphics[width=0.70\columnwidth]{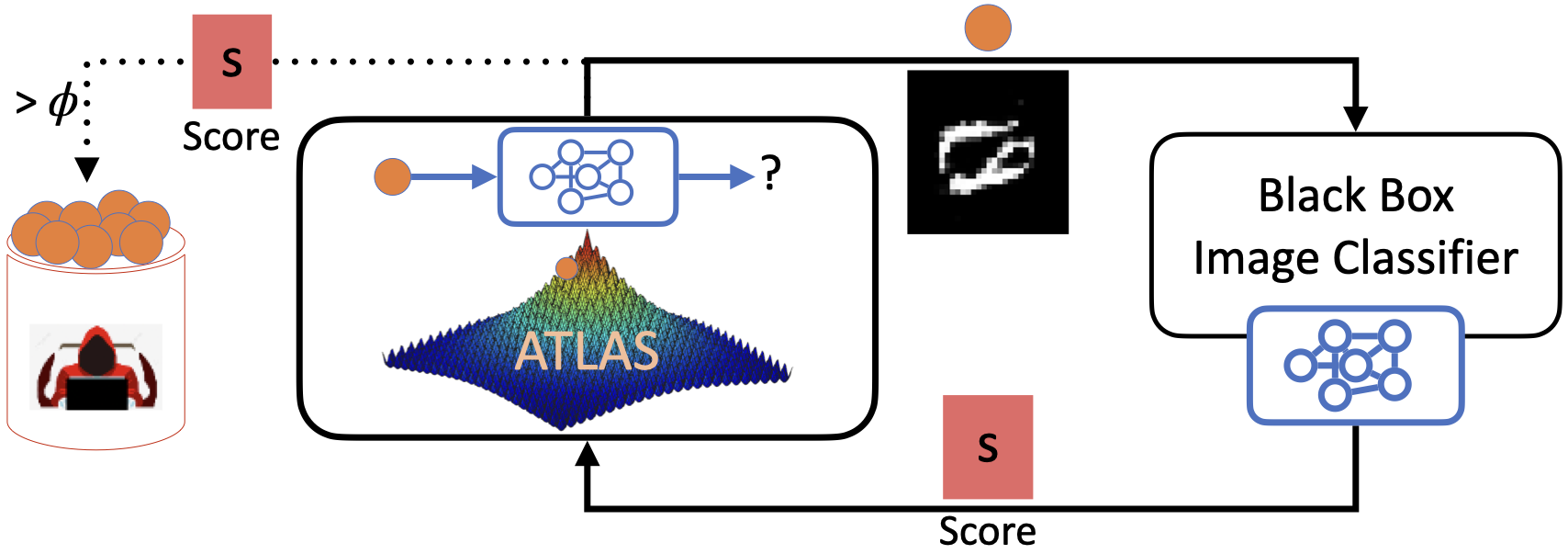} 
    \caption{\textbf{ATLAS}. Given a target (black box image classifier), ATLAS searches for adversarial images (orange dot) by learning a mapping between inputs (orange dot) and scores that represent the level of adversity of those images for the target model. Using this function, ATLAS produces a set of attacks (the red bucket) for the target model.}
    \label{fig:method_summary}
\end{figure}

As more learning-based systems get deployed in safety critical settings \citep{eykholt2018robust, chou2021reachability}, robust evaluation requires more than testing against a fixed collection of previously known attacks\citep{athalye2018obfuscated, croce2020reliable}. An evaluator must be able to adaptively discover diverse inputs that induce failure, especially for complex targets whose failure mechanisms are unknown \citep{nasr2025attacker, bahati2021multi}. Existing black-box adversarial attacks are designed to discover one/few successful failures. However, in our problem, our goal is to identify a set of adversarial inputs that produces failures for a specified target system. Rather than optimizing for a single successful adversarial example, this formulation seeks to produce a wide-spread adversarial set that can be used to audit where and how a learning-based system fails (Figure~\ref{fig:method_summary}). The challenge comes from the input space being high-dimensional, making exhaustive search infeasible, and potentially having input adversarial regions occupy only a small fraction of the overall search space, making unguided exploration unlikely to discover them. 

Therefore, we introduce \textsc{ATLAS}, an active-learning framework for generating adversarial sets in high-dimensional spaces. One of the key innovations is that it \textit{re-formulates adversarial discovery as a region-discovery problem and combines tools from level-set estimation, Bayesian optimization, and adversarial generation to build these sets}. Because existing approaches \citep{ngo2026high, eriksson2019scalable, deglurkar2026active, paccagnan2023pick} are not enough to solve this problem -  these approaches aren't designed to generate diverse adversarial sets - ATLAS builds upon previous work and uses calibrated model predictions combined with a dual sampling mechanism to build both an adversarial classifier and the associated adversarial dataset. To evaluate ATLAS, we first use non-convex benchmark functions helps us analyze ATLAS' ability to recover sparse and geometrically complex adversarial sets in toy problems. After this, we apply ATLAS on several image classification models. Since existing automated attack frameworks primarily uses optimization-based approaches, only finding single points of failure at a time, we investigate whether or not the repeated application of existing attack-generation methods is enough to recover diverse adversarial sets or whether a method tailored to adversarial set construction would provide some advantage. When compared against popular optimization-based methods, we find that not only is ATLAS able to recover sets with more empirical coverage of the adversarial region, but it is also able to curate attacks with a success rate that is on par or better than the baselines. 

\textbf{Our contributions are}: (1) framing adversarial generation as an active level-set estimation problem rather than an optimization problem, and showing that optimization-based attacks recover less of the adversarial set even when all of their queries are retained; (2) calibrating the surrogate's predictions with a residual model trained over the entire input space, in contrast to prior work that calibrates the sample-selection score; (3) designing a saturation trust region and acquisition that recover the interior of the adversarial set rather than only its boundary; and (4) evaluating on non-convex benchmarks with known ground-truth level sets up to 100D, and on standard and adversarially trained MNIST, CIFAR-10, and ImageNet classifiers, with ablations isolating (2) and (3). 


\section{Related Works}
\label{sec:related-works}
There are certain categories of methods that generate image attacks: (a) Methods that use gradient-based approaches to craft adversarial inputs \citep{goodfellow2014explaining,eykholt2018robust,papernot2017practical,kurakin2018adversarial,carlini2017towards,ilyas2019adversarial,athalye2018obfuscated,kurakin2016adversarial,croce2020reliable}, (b) methods that use generative models to generate adversarial images \citep{xiao2018generating,jandial2019advgan++,song2018constructing}, \citep{poursaeed2018generative,xu2023dcvae}, and (c) query-based methods that finds attacks to black box systems. In the latter, there exist Bayesian Optimization based approaches \citep{shukla2019black, cheng2024efficient,zhang2024black,ru2019bayesopt} and generic querying methods like SquareAttack \citep{andriushchenko2020square}, SignHunter \citep{al2020sign}, and NES \citep{ilyas2018black}. Most notably, BayesOpt \citep{ru2019bayesopt} is a well known method that is most similar to ours. In this work, they aim to find $L_{\infty}$ based perturbation using similar adaptive sampling mechanisms as we do; however, they get around the curse of dimensionality by decomposing the input space into lower dimensional subspaces and assigning a local model to represent the adversarial objective in that lower-dimensional space. This is complementary to our method. In ATLAS, our input space is decomposed into local regions of the same dimension, while BayesOpt decomposes its search space into manifolds of lower dimensions. Furthermore, their method is designed to optimize an adversarial objective to find one specific attack. Our framework is designed to learn an adversarial mapping globally; thereby learning entire sets of adversarial examples that exists for a target system. Because of the differing aims, our adaptive sampling mechanisms are geared towards minimizing model error vs. generic adversarial maximization. Other approaches leverage LM-evolutionary methods \citep{nasr2025attacker}, \citep{zhan2025adaptive}, and generic try-and-check methods \citep{gibson2024hacking} as well.

\section{Threat Model and Problem Formulation}
\label{Sec: Threat Model}
Let $\targetsystem: \inputspace \rightarrow \targetoutputspace$ represent a black box image classification system that maps from a bounded image space $\inputspace$ to a probability distribution representing its output classification $\targetoutputspace$. Let $\adversarialobjective: \targetoutputspace \rightarrow \adversarialscore$ represent our given adversarial objective that scores the output of the target system with respect to some adversarial criteria and let a successful attack be characterized by $\adversarialscore > \threshold$. Also, assume that the attack framework has limited query access to the target system $\targetsystem$ and can observe output probabilities. We are interested in producing a set of adversarial inputs, $\genericadversarialset \coloneqq \{\inputsample \in \inputspace: \advfunclandscape(\inputsample) \geq \threshold \}$, that represents a set of images that can successfully attack the system $\advfunclandscape$, where $\advfunclandscape(\inputsample) = (\targetsystem \circ \adversarialobjective)(\inputsample)$. To generate this adversarial set, we formulate a level-set estimation active learning problem, where we aim to learn an adversarial value function $\valuefunc(\inputsample) \rightarrow \adversarialscore$ that maps candidate inputs $\inputsample$ to adversarial scores $\adversarialscore \in \mathbf{R}$, whose super-zero level set $\superlevelset \coloneqq \{\inputsample : \valuefunc(\inputsample) \geq 0\}$ is the adversarial set we aim to recover. Previous works \cite{ngo2026high, deglurkar2026active} solves this problem by sampling near the zero level set. While sufficient for classification (are points safe or adversarial), this leaves the interior of $\superlevelset$ under-sampled which is unfavorable since an audit also requires a representative sample of $\superlevelset$ itself. However, trying to sample the interior of $\superlevelset$ is not trivial. Interior candidates can only be chosen from the surrogate's own prediction of where $\superlevelset$ lies. This prediction is often wrong and causes queries to land outside of $\superlevelset$ or in already highly sampled regions the model knows well. While prior work \citep{deglurkar2026active} attempt to calibrate where samples are selected using calibrated acquisition functions, these functions combine prediction and uncertainty on an arbitrary scale, which makes it both harder to calibrate and redundant to do so: the calibration must be repeated for every acquisition function, and none of it corrects the underlying model. Our key idea is therefore to calibrate the model's prediction directly, using a residual model trained on the main model's error over the entire input space. A single correction then propagates to every acquisition built on that model. This lets us define two complementary selection rankings over a shared calibrated surrogate: one preferring candidates near the zero level set (\emph{boundary}), which improves classification of the adversarial region, and one preferring candidates predicted to lie inside $\superlevelset$ (\emph{saturation}), which curates the adversarial set.

\section{ATLAS: Adaptive Trust-Region for Latent Adversarial Search}
\label{Sec: Methodology}
\begin{figure}[t] 
    \centering
    \includegraphics[width=0.70\columnwidth]{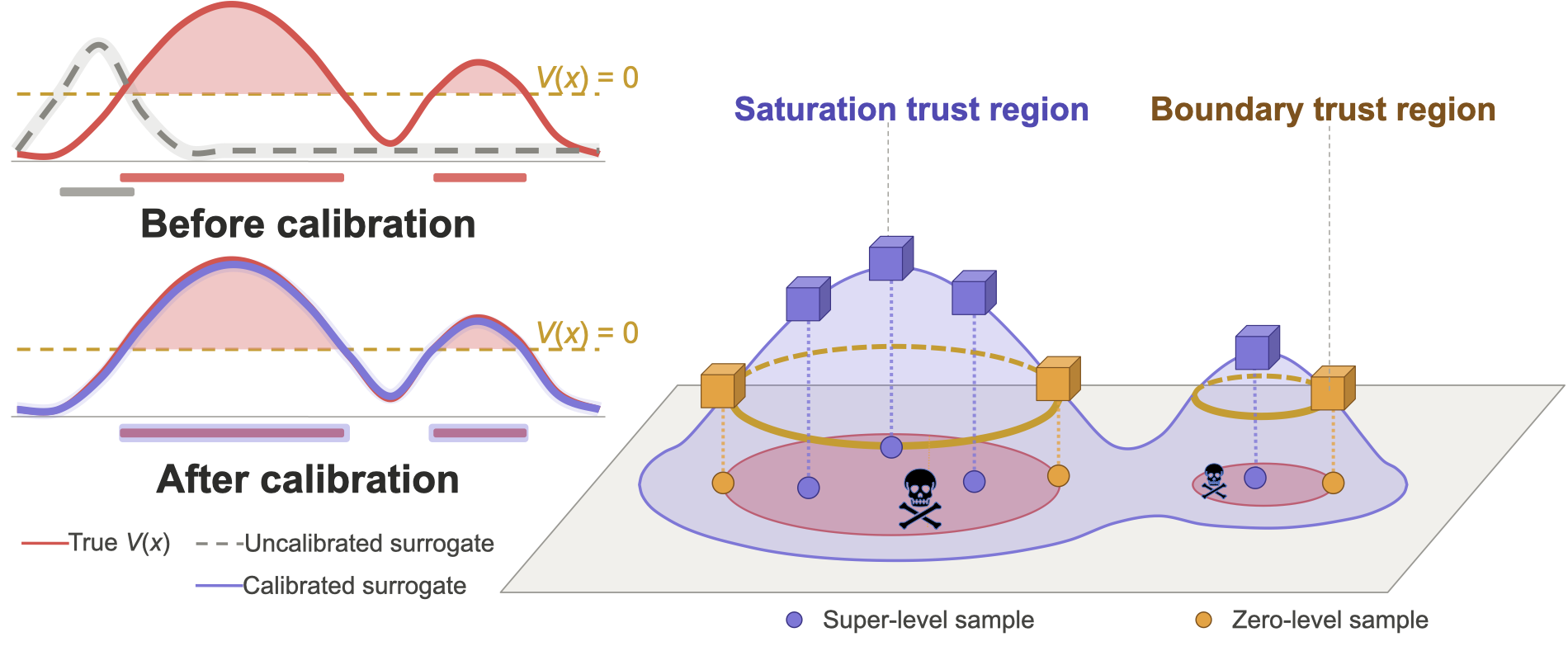} 
    \caption{ATLAS performs level set estimation to find regions containing adversarial examples. It queries the target system using calibrated model predictions (purple surface) and a dual sampling strategy. The first sample points at the zero level set (gold boxes) to find the edges of adversarial regions, while the other sample within the super level set (purple boxes) to collect as many adversarial examples as possible. ATLAS produces both a model that correctly estimates which inputs are adversarial and an adversarial set that provides good coverage of target's adversarial region.}
    \label{fig:method_overview}
\end{figure}

ATLAS (Fig. \ref{fig:method_overview}) constructs both an adversarial set $\genericadversarialset$ and an adversarial value function $\valuefunc(\inputsample) \rightarrow \adversarialscore$ using two components: (i) a calibration scheme for the trained value function and (ii) a local placement-sampling scheme that samples from two types of local regions. The first type (boundary regions) focuses on sampling from the zero level set for discovery, and the other (saturation regions) samples within discovered super level sets to produce $\genericadversarialset$. We refer to these local regions as `trust regions'. A single algorithm, Algorithm \ref{alg:ATLAS}, ties these two together: (1) fit global model, (2) fit residual model and calibrate, (3) select trust regions, (4) sample locally, and (5) refit with new local samples and repeat.

\begin{algorithm}[ht]
\DontPrintSemicolon
\SetKwInOut{Input}{Input}
\SetKwInOut{Initialize}{Initialize}
\SetKwComment{Comment}{$\triangleright$\ }{ }
\SetKwFunction{LocalSample}{LocalSample}
\SetAlgorithmName{Algorithm}{algorithm}{List of Algorithms}
\caption{ATLAS: Adaptive Trust-Region for Latent Adversarial Search}
\label{alg:ATLAS}

\Input{%
  $\locallabelingfunc$ (oracle); \newline
  $\numboundaryregions, \numsatregions$ (number of boundary / saturation regions); \newline
  $\localsamplingbudget$ (queries per region per round); \newline
  $\globalsamplingbudget$ (total query budget)
}
\Initialize{%
  $\calibration \gets \{(\inputsample_i, \locallabelingfunc(\inputsample_i))\}_{i=1}^{\certificationsetsize},\ \inputsample_i \sim \mathcal{U}(\inputspace)$; (held out; for residual models only)\newline
  $\datasetsamples = \emptyset$; $t = 0$
}
\BlankLine
\While{$|\datasetsamples| + |\calibration| < \globalsamplingbudget$}{
  \Comment{1: Fit the global surrogate}
  $\globalmodel \gets \learningalgorithm(\datasetsamples)$ \tcp*[r]{Models $\valuefunc$}\;
  \BlankLine
  
  \Comment{2: Fit a residual model and calibration global surrogate}
  $\residualmodel \gets \learningalgorithm\Big(\{ (\inputsample_i,\, \globalmodel(\inputsample_i) - \locallabelingfunc(\inputsample_i)) \mid \inputsample_i \in \calibration \}\Big)$\;
  $\calibratedmodel \gets \globalmodel - \residualmodel$  
  \BlankLine
  
  \Comment{3: Choose boundary and saturation regions centers}
    \For{$k \gets 1$ \KwTo $\numboundaryregions$}{
        $\seed_k \gets \arg \max_{\inputsample \in \inputspace} \boundaryglobalacquisition(\inputsample, \calibratedmodel)\;+\;\penaltyfuncweight\!\prod_{\seed_j \in \boundtrustregioncenters^t}\! \penaltyfunc\bigl(\lVert \datasetsample-\seed_j\rVert_2^2\bigr)$\;
        $\boundtrustregioncenters^t \gets \boundtrustregioncenters^t \cup \{\seed_k\}$\;
    }
    \For{$k \gets 1$ \KwTo $\numsatregions$}{
        $\seed_k \gets \arg \max_{\inputsample \in \inputspace} \satglobalacquisition(\inputsample, \calibratedmodel)\;+\;\penaltyfuncweight\!\prod_{\seed_j \in \sattrustregioncenters^t}\!
      \penaltyfunc\bigl(\lVert \datasetsample-\seed_j\rVert_2^2\bigr)$\;
      $\sattrustregioncenters^t \gets \sattrustregioncenters^t \cup \{\seed_k\}$\;
    }
  \BlankLine
  
  \Comment{4: Sample locally in every region}
  $\compressionset \gets \LocalSample(\boundtrustregioncenters^t, \hat a_B, \datasetsamples) \bigcup \LocalSample(\sattrustregioncenters^t, \hat a_S, \datasetsamples)$\;
  \BlankLine
  
  \Comment{5: Add the new samples; the next iteration refits}
  $\datasetsamples \gets \datasetsamples \cup \compressionset$; \quad $t \gets t + 1$; \quad $\boundtrustregioncenters^t = \emptyset$; \quad $\sattrustregioncenters^t = \emptyset$\; 
}
\Return calibrated surrogate $\calibratedmodel$;\quad adversarial set $\{\inputsample \in \datasetsamples : \locallabelingfunc(\inputsample) \ge \threshold\}$
\end{algorithm}

\begin{algorithm}[ht]
\DontPrintSemicolon
\SetKwInOut{Input}{Input}
\SetKwComment{Comment}{$\triangleright$\ }{ }
\SetAlgorithmName{Algorithm}{algorithm}{List of Algorithms}
\caption{LocalSample: Calibrated Sampling inside Trust Regions}
\label{alg:local}

\Input{Trust Region Centers $\trustregioncenters$; local acquisition $\genericlocalacquisition$ ($\boundarylocalacquisition$ or $\satlocalacquisition$); samples $\datasetsamples$; local sampling budget $\localsamplingbudget$}

\BlankLine
regions $\trustregion_i \gets \trustregioncenters_i$; \quad
$\compressionset \gets \emptyset$;\quad $\unlabeleddataset_{\trustregion} \gets$ candidates drawn in $\trustregion$\;
\If{$\datasetsamples \cap \trustregion = \emptyset$}{
  $\compressionset \gets \{(\trustregioncenter,\, \locallabelingfunc(\trustregioncenter))\}$ \tcp*[r]{empty region: label its center}
}
\For{$m = 1, \dots, \localsamplingbudget$}{
  \Comment{Calibrate local model by LOO-CV \citep{williams2006gaussian}}
  $\localcalibration \gets (\datasetsamples \cap \trustregion) \cup \compressionset$\;
  $\localcalibratedmodel \gets \learningalgorithm(\localcalibration) - \learningalgorithm\Big(\big\{ \big(\inputsample_i,\; h^{(-i)}(\inputsample_i) - \locallabelingfunc(\inputsample_i)\big) \mid \inputsample_i \in \localcalibration \big\}\Big)$, \quad $h^{(-i)} = \learningalgorithm(\localcalibration \setminus \{\inputsample_i\})$\;
  \BlankLine
  \Comment{take the best candidate under the calibrated local model}
  $\inputsample^\star \gets \arg\max_{\inputsample \in \unlabeleddataset_{\trustregion} \setminus \compressionset}\ \hat a(\inputsample;\ \localcalibratedmodel)$\;
  $\compressionset \gets \compressionset \cup \{(\inputsample^\star, \locallabelingfunc(\inputsample^\star))\}$\;
}
\Return $\compressionset$
\end{algorithm}

\subsection{Calibration}
Calibration helps turn inaccurate model estimates to accurate model estimates, helping us to efficiently leverage our sampling budget. To successfully calibrate our model, we can take a hold out set from our budget, train a second model that estimates the main model's error per sample (its residual), and apply the estimate error as a residual offset on our main model's prediction. However, in our application because each local region contains its own local model, we must calibrate both the global model for accurate placement, and our local model for accurate sampling. This is challenging because every local region is temporary and moves. Successfully calibrating each and every local model with a fresh hold out set as they get reinitialized in a new part of the space will easily eat up our sampling budget. Therefore, we devise two separate calibration schemes - one for our local models and another for the global model. Our global model's calibration scheme uses a uniformly sampled held out set to train its residual model (see Algorithm \ref{alg:ATLAS}, while our local models use leave-one-out cross-validation (LOO-CV) from \citep{williams2006gaussian} using only its training points to avoid depleting the sampling budget.

\subsection{Acquisition Design for Placement and Sampling}
ATLAS follows the classic high dimensional global-local sampling architecture from previous works \citep{ngo2026high, eriksson2019scalable}, where, first, local trust regions are selected, and then, points are sampled locally within those regions. As a consequence, there are two sampling loops - an outer loop that select trust region centers $\trustregioncenters^{t}$ for iteration $t$ and inner loop that selects the next best samples for our dataset. Let $\trustregion = \{ \datasetsample \in \mathbb{R}^\trustregionsampledimension \mid \Vert{} \datasetsample - \trustregioncenter \Vert{}_\infty < \frac{\trustregionlength}{2}\}$ denote our trust regions, which are hypercubes centered at $\trustregioncenter$ with side length $\trustregionlength$. Each trust region's center $\seed_i$ is chosen as 
\begin{equation}
\begin{aligned}
\seed_{j+1} &\;=\;\operatorname*{arg\,max}_{\inputsample \in \mathcal{U}(\inputspace) \setminus \trustregioncenters^{t-1}}\;
   \genericglobalacquisition(\datasetsample)\;+\;\penaltyfuncweight\!\!\prod_{\seed_j \in \seedset}\!
   \penaltyfunc\bigl(\lVert \datasetsample-\seed_j\rVert_2\bigr)
\end{aligned}
\label{eq:place}
\end{equation}
where $\penaltyfunc(\penaltyfuncinput) \;=\;\frac{1}{1+e^{-(\penaltyfuncinput-\seedseparation)/(\penaltyfuncweight*\seedseparation)}}$ represents a distance penalty, parameters $\seedseparation$ and $\penaltyfuncweight$ represent desired seed separation and a multiplier respectively, $\genericglobalacquisition$ represents a designed acquisition placement function, and trust region centers $\trustregioncenters^{t-1}$ are the centers previously chosen during the last iteration. Every sample is selected simply using a designed acquisition sampling function $\alpha$. Designing these acquisition functions is not trivial. Because ATLAS learns both an adversarial value function while curating an adversarial dataset, each acquisition must be designed in a way that improves the model's classification and increases the diversity of the adversarial set. This is challenging because each design criterion is not necessarily aligned. Prioritizing model classification often means training on points with higher model error, whereas sampling diverse adversarial points may mean sampling from regions you already know. To solve this problem, we devise two different acquisition functions: one that selects zero level samples with high model error and another that selects super level samples with high model error. We denote these zero-level based (boundary) and super-level based (saturation) placement acquisitions with $\boundaryglobalacquisition$ and $\satglobalacquisition$ and their corresponding sampling functions with $\boundarylocalacquisition$ and $\satlocalacquisition$. Let the Gaussian Process $\localgp{i}$ with posterior mean $\localmean(x)$ and variance $(\localuncertainty(x))^2$ represent our calibrated model $\genericcalibratedmodel$. Our boundary-focused and saturation-focused sampling function selects candidates based on:
\begin{equation}
\begin{aligned}
    \boundarylocalacquisition(\inputsample, \localmean, \localuncertainty)\;=\;
    (1-\localboundarymixingweight) e^{-\bigl|\localmean(\inputsample)\bigr|/s_{\mathrm{bnd}}}
    + \localboundarymixingweight(1- e^{-\boundaryweight\beta\localuncertainty(\inputsample)/s_{\mathrm{bnd}}})
\end{aligned}
\label{eq:a-boundary}
\end{equation}
and 
\begin{equation}
\begin{aligned}
    \satlocalacquisition(\inputsample, \localmean, \localuncertainty)\;=\;
    1-e^{-\left(\beta\,\localuncertainty(x) + \superlevelprobweight\, \postprob \right) /s_{\mathrm{sat}}}
\end{aligned}
\label{eq:a-saturation}
\end{equation}
where 
\begin{equation}
\begin{aligned}
    P\ =
    p_s(\inputsample, \localmean(\inputsample), \localuncertainty(\inputsample))\;=\;1-\Phi\!\left(\frac{-\localmean(x)}{\localuncertainty(x)}\right)
\end{aligned}
\label{eq:post_prob}
\end{equation}
denotes the posterior probability that $\inputsample$ belongs to the adversarial super-level set. $\Phi$ represents the CDF of a normal distribution. Boundary (leveraging  \cite{bryan2005active}) and saturation placements are designed as:
\begin{equation}
\begin{aligned}
\boundaryglobalacquisition(\inputsample) = \boundaryweight\,\globaluncertaintyparam\,\localuncertainty(\inputsample)\ -  \bigl|\localmean(\inputsample)\bigr|
\end{aligned}
\label{eq:global-bound-acq}
\end{equation}
and
\begin{equation}
\begin{aligned}
\satglobalacquisition(\inputsample) = \globaluncertaintyparam\,\localuncertainty(\inputsample)\ + 
\globaluncertaintyparam\,\localuncertainty(\inputsample)\,\postprob\,(1-\postprob)\ +
\superlevelprobweight\,\postprob\
\end{aligned}
\label{eq:global-sat-acq}
\end{equation}
Each trust region is a hypercube of fixed volume $\volume$, i.e., side length $\trustregionlength = \volume^{1/\dimension}$ (Table~\ref{tab:hyperparams}).  
\section{Toy Experiments: Finding Adversarial Regions in Highly Non-Convex Functions}
\label{sec: toy_exp}

We design a controlled experiment in which our black-box adversarial target $\advfunclandscape$ is represented by a non-convex function $\toyfunction$ (see Appendix \ref{Appendix: Function}) and the goal is to recover a super-level set representing the space of possible attacks for the system $\genericadversarialset \coloneqq \{\toyinput \in \mathrm{R}^\toyinputdim: \toyfunction(\toyinput) > \threshold \}$. We compare ATLAS to AP2L \citep{deglurkar2026active} and TRLSE \citep{ngo2026high}. We use 2 test sets consisting of 50K super level samples ($\superlevelset$) and 1 million mixed points (Mix), where the mixed test set contains all super level samples and 950K other uniformly sampled points. A threshold $\threshold = \operatorname{Quantile}\!\left( \left\{ \advfunclandscape(\inputsample_i) \right\}_{i=1}^{\thresholdsetsize}, \, 1-\superlevelratio \right)$ for each experiment is chosen such that the ratio of the volume of the super level set to the volume of the total domain matches a chosen parameter $\superlevelratio = 0.05$, where $\inputsample \sim \mathcal{U}(\inputspace)$. For model performance, we measure the average F1 score and report recall (over $\superlevelset \subset \text{Mix}$) and precision. For each adversarial set, we measure the spread of adversarial samples using the mean distance between each pair of samples in the set: 
\begin{equation}
    \dispersionmetric(\genericadversarialset) = \binom{M}{2}^{-1} \sum_{i \neq j} \Vert{}\genericadversarialset_i - \genericadversarialset_j\Vert{}_2
\end{equation}
where $M = |\genericadversarialset|$. This metric is normalized by the spread of the test set for each experiment $\dispersionmetricnorm(\genericadversarialset) = \frac{\dispersionmetric(\genericadversarialset)}{\dispersionmetric(\superlevelset)}$. Because this metric biases towards sets containing boundary samples (samples at the edge of the super level set will by definition have higher average pairwise distance than samples spread within the interior), we include 2 extra metrics that measures how spread the samples' adversarial scores are within the set and how deep inside the interior each method is able to go. The spread of adversarial scores (Margin J) is measured across $K=20$ discretized bins $b_k$ over the interval $[0, f(\inputsample^*) - \threshold]$:
\begin{equation}
    J(\genericadversarialset) = -\frac{1}{\ln K} \sum_{k=1}^{K} p_k \ln p_k, \qquad p_k = \frac{|\{x \in \genericadversarialset : f(x) - \threshold \in b_k\}|}{|\genericadversarialset|}
\end{equation}
where $J(\genericadversarialset) \in [0, 1]$ measures the normalized Shannon entropy ($1 = \text{uniform coverage}$, $0 = \text{single bin concentration}$). Super level penetration is measured using the maximum adversarial score contained in each adversarial set divided by the largest possible adversarial score: 
\begin{equation}
    \text{mV}(\genericadversarialset) = \max_{\inputsample \in \genericadversarialset} \frac{f(\inputsample) - \threshold}{f(\inputsample^*) - \threshold} 
\end{equation} This metric is reported alongside with the $\%$ of samples inside the adversarial set that exceeds the test set's ($\superlevelset$) largest adversarial score. The following tables report two ablation experiments and our main benchmark experiment. Table~\ref{tab:calibration-ablation-paper} and Table~\ref{tab:saturation-ablation-paper} report representative ablation results across all toy functions in $100\text{D}$, while Table~\ref{tab:joint-eval-multi-dim-main} contains the full results of our benchmark experiment; all extended analysis is in Appendix~\ref{sec: appendix-ext}. All experiments use $\numseeds$ seeds and a budget of $\expbudget$, with the best-performing calibration budget from our sweep ($17\%$, Table~\ref{tab:calibration-ablation-appendix}). Calibration queries count against ATLAS's budget; AP2L receives its calibration set outside its budget. Full settings for ATLAS and the baselines are in Table~\ref{tab:hyperparams}, and Table~\ref{tab:default-table} reports ATLAS with a single default weight setting shared across all benchmarks.


\begin{table}[htbp]
\centering
\caption{\textbf{Calibration is what makes ATLAS work in high dimensions.} We vary the number of calibration samples $N_C$ drawn from the fixed total budget $B = 650$, leaving $B - N_C$ for sampling. We use 30 seeds. At 100D, adding just 50 calibration samples restores classification performance (from .163, .003, and .100 to .322, .288 and .457 for Ack, Rast, Rosen). The benefit grows with dimension (See table~\ref{tab:calibration-ablation-appendix} for full analysis).}
\label{tab:calibration-ablation-paper}
\begin{threeparttable}
\scriptsize
\setlength{\tabcolsep}{5pt}
\begin{tabular}{lll ccc}
\toprule
\textbf{Func} & \textbf{Dim} & \textbf{Calib. $N_C$ (\% of B)} & \textbf{F1 (Mix) $\uparrow$} & \textbf{Rec $\superlevelset$ \% $\uparrow$} & \textbf{Prec (Mix) \% $\uparrow$}\tnote{1} \\
\midrule

\multirow{5}{*}{\textbf{Ackley}} 
  & \multirow{5}{*}{100D}
    & $0$ (0\%) & $0.163_{.020}$ & $9.2_{1.3}$ & $\mathbf{77.1}_{2.7}$ \\
  & & $50$ (8\%) & $\mathbf{0.322}_{.069}$ & $\mathbf{22.8}_{6.2}$ & $60.3_{6.3}$ \\
  & & $100$ (15\%) & $0.305_{.056}$ & $21.0_{5.7}$ & $60.6_{5.1}$ \\
  & & $250$ (38\%) & $0.199_{.050}$ & $12.1_{3.7}$ & $63.4_{5.4}$ \\
  & & $375$ (58\%) & $0.029_{.028}$ & $1.5_{1.5}$ & $72.9_{10.6}$ \\
\midrule

\multirow{5}{*}{\textbf{Rast.}} 
  & \multirow{5}{*}{100D}
    & $0$ (0\%) & $0.003_{.002}$ & $0.2_{0.1}$ & $\mathbf{94.6}_{3.2}$ \\
  & & $50$ (8\%) & $\mathbf{0.288}_{.051}$ & $\mathbf{20.3}_{5.2}$ & $54.0_{6.1}$ \\
  & & $100$ (15\%) & $0.278_{.035}$ & $19.1_{3.5}$ & $53.7_{4.3}$ \\
  & & $250$ (38\%) & $0.251_{.030}$ & $16.8_{2.7}$ & $51.0_{3.3}$ \\
  & & $375$ (58\%) & $0.214_{.046}$ & $13.7_{3.8}$ & $51.6_{4.1}$ \\
\midrule

\multirow{5}{*}{\textbf{Rosen.}} 
  & \multirow{5}{*}{100D}
    & $0$ (0\%) & $0.100_{.035}$ & $5.3_{2.0}$ & $\mathbf{90.2}_{1.6}$ \\
  & & $50$ (8\%) & $\mathbf{0.457}_{.037}$ & $\mathbf{43.6}_{9.6}$ & $52.2_{9.9}$ \\
  & & $100$ (15\%) & $0.455_{.041}$ & $41.9_{8.0}$ & $52.8_{8.9}$ \\
  & & $250$ (38\%) & $0.447_{.026}$ & $39.0_{4.6}$ & $53.9_{7.1}$ \\
  & & $375$ (58\%) & $0.423_{.025}$ & $36.1_{4.1}$ & $52.7_{7.5}$ \\
\bottomrule
\end{tabular}
\begin{tablenotes}
\item[1] No calibration attains high precision by predicting few super-level points, at the cost of recall. This is undesirable for a security analysis tool, whose purpose is to catch as many adversarial inputs as possible.
\end{tablenotes}
\end{threeparttable}
\end{table}

\begin{table}[t]
\begin{threeparttable}
\centering
\caption{\textbf{In 100D, saturation and boundary sampling allow ATLAS to produce a more diverse adversarial set while having robust model estimation.} 
All variants share the same sampling budget (B = 538) and number of trust regions; ATLAS-b uses only boundary regions, ATLAS-s only saturation, and ATLAS a 1:1 split. Subscripts are standard deviations. Each single-mode variant performs as designed. Saturation sampling harvests adversarial points (nSuper/B $\approx$ 90\%), while boundary sampling gives the widest spatial spread. Each fails somewhere: ATLAS-s collapses in F1 on Rastrigin and Rosenbrock ($\approx 0.01$), and ATLAS-b finds almost no adversarial points on Rastrigin. Only ATLAS is robust across all landscapes. See Table \ref{tab:saturation-ablation-appendix} in the Appendix for full analysis.}
\label{tab:saturation-ablation-paper}
\scriptsize
\setlength{\tabcolsep}{2pt}
\begin{tabular}{lll ccc c ccccc}
\toprule
& & & \multicolumn{3}{c}{\textbf{Learned Model Metrics}} & & \multicolumn{5}{c}{\textbf{Sampling Set Metrics}} \\
\cmidrule(lr){4-6} \cmidrule(lr){8-12}
\textbf{Func} & \textbf{Dim} & \textbf{Arm} & \textbf{F1 (Mix) $\uparrow$} & \textbf{Rec $\superlevelset$ \% $\uparrow$} & \textbf{Prec (Mix) \% $\uparrow$} & & \textbf{mV ($\%$) $\uparrow$} & \textbf{Margin J $\uparrow$} & \textbf{$\dispersionmetricnorm$ $\uparrow$} & \textbf{nSuper/B (\%) $\uparrow$} & \textbf{B} \\
\midrule

\multirow{3}{*}{\textbf{Ackley}} 
  & \multirow{3}{*}{100D}
    & \textsf{ATLAS-b} & $0.223_{.082}$ & $14.0_{6.0}$ & $\mathbf{68.9}_{9.1}$ & & $0.039_{.022}\,(1.9)$ & $0.001_{.004}$ & $\mathbf{0.913}_{.026}$ & $48.3_{6.5}$ & 538 \\
  & & \textsf{ATLAS-s} & $\mathbf{0.311}_{.067}$ & $\mathbf{21.6}_{6.4}$ & $60.4_{4.8}$ & & $\mathbf{0.053}_{.005}\,(21.9)$ & $\mathbf{0.006}_{.005}$ & $0.825_{.010}$ & $\mathbf{85.7}_{2.8}$ & 538 \\
  & & \textsf{ATLAS}   & $0.307_{.038}$ & $20.8_{3.6}$ & $60.6_{3.7}$ & & $0.051_{.006}\,(20.3)$ & $\mathbf{0.006}_{.007}$ & $0.837_{.007}$ & $52.0_{2.2}$ & 538 \\
\midrule

\multirow{3}{*}{\textbf{Rast.}} 
  & \multirow{3}{*}{100D}
    & \textsf{ATLAS-b} & $0.255_{.059}$ & $16.2_{5.1}$ & $67.1_{5.7}$ & & $0.119_{.063}\,(0.0)$ & $0.249_{.128}$ & $0.816_{.143}$ & $1.5_{1.1}$ & 538 \\
  & & \textsf{ATLAS-s} & $0.011_{.010}$ & $0.6_{0.5}$ & $\mathbf{83.2}_{8.1}$ & & $\mathbf{0.317}_{.017}\,(4.2)$ & $\mathbf{0.532}_{.013}$ & $\mathbf{1.080}_{.021}$ & $\mathbf{90.9}_{2.0}$ & 538 \\
  & & \textsf{ATLAS}   & $\mathbf{0.273}_{.036}$ & $\mathbf{18.7}_{3.8}$ & $52.9_{3.8}$ & & $0.304_{.023}\,(3.6)$ & $0.515_{.017}$ & $1.032_{.018}$ & $50.0_{1.7}$ & 538 \\
\midrule

\multirow{3}{*}{\textbf{Rosen.}} 
  & \multirow{3}{*}{100D}
    & \textsf{ATLAS-b} & $\mathbf{0.503}_{.052}$ & $\mathbf{62.5}_{11.3}$ & $45.2_{10.9}$ & & $0.586_{.150}\,(3.8)$ & $0.649_{.080}$ & $\mathbf{0.959}_{.027}$ & $22.3_{8.6}$ & 538 \\
  & & \textsf{ATLAS-s} & $0.015_{.011}$ & $0.8_{0.6}$ & $\mathbf{94.9}_{3.0}$ & & $0.657_{.030}\,(21.2)$ & $0.786_{.011}$ & $0.799_{.007}$ & $\mathbf{90.9}_{1.7}$ & 538 \\
  & & \textsf{ATLAS}   & $0.461_{.028}$ & $41.0_{5.7}$ & $55.0_{8.4}$ & & $\mathbf{0.844}_{.028}\,(32.8)$ & $\mathbf{0.914}_{.011}$ & $0.839_{.007}$ & $47.6_{2.8}$ & 538 \\
\bottomrule
\end{tabular}
\end{threeparttable}
\end{table}

\begin{table}[t]
\centering
\caption{\textbf{ATLAS produces better recall and F1 while recovering more adversarial points per query, using a smaller sampling budget (B).} We compare ATLAS's calibration (112 samples) and 1:1 boundary-to-saturation sampling against our baselines at $B = 538$ vs.\ 650. Subscripts are standard deviations. ATLAS trades precision for recall (recall best in 9 of 9 and F1 (Mix) best in 8 of 9), yielding a model that misses fewer adversarial inputs. It also finds the most super-level points per unit budget (8 of 9) and recovers a better-distributed set — the highest margin spread and depth in every experiment, with dispersion within 30\% of the best.\tnote{1}}
\label{tab:joint-eval-multi-dim-main}
\begin{threeparttable}
\scriptsize
\setlength{\tabcolsep}{1.5pt}
\begin{tabular}{lll ccc c ccccc}
\toprule
& & & \multicolumn{3}{c}{\textbf{Learned Model Metrics}} & & \multicolumn{5}{c}{\textbf{Sampling Set Metrics}} \\
\cmidrule(lr){4-6} \cmidrule(lr){8-12}
\textbf{Func} & \textbf{Dim} & \textbf{Method} & \textbf{F1 (Mix) $\uparrow$} & \textbf{Rec $\superlevelset$ \% $\uparrow$} & \textbf{Prec\% $\uparrow$} & & \textbf{mV ($\%$) $\uparrow$} & \textbf{Margin J $\uparrow$} & \textbf{$\dispersionmetricnorm$ $\uparrow$} & \textbf{nSuper/B (\%) $\uparrow$} & \textbf{B} \\
\midrule

\multirow{9}{*}{\textbf{Ackley}} 
  & \multirow{3}{*}{10D}
    & AP2L   & $0.088_{.203}$ & $14.0_{29.9}$ & $49.3_{37.9}$ & & $0.096_{.033}\,(0.0)$ & $0.128_{.072}$ & $0.957_{.102}$ & $5.5_{9.1}$ & 650 \\
  & & TRLSE  & $\mathbf{0.681}_{.005}$ & $59.2_{1.5}$ & $\mathbf{80.2}_{1.5}$ & & $0.113_{.044}\,(0.0)$ & $0.042_{.011}$ & $\mathbf{1.019}_{.005}$ & $26.0_{2.0}$ & 650 \\
  & & ATLAS  & $0.664_{.020}$ & $\mathbf{63.6}_{4.9}$ & $69.8_{4.1}$ & & $\mathbf{0.304}_{.048}\,(0.0)$ & $\mathbf{0.368}_{.016}$ & $0.735_{.033}$ & $\mathbf{53.7}_{3.3}$ & 538 \\
  \cmidrule(lr){2-12}

  & \multirow{3}{*}{50D}
    & AP2L   & $0.010_{.024}$ & $0.5_{1.3}$ & $\mathbf{81.6}_{4.5}$ & & $0.021_{.009}\,(0.0)$ & $0.000_{.000}$ & $\mathbf{0.991}_{.009}$ & $2.9_{3.5}$ & 650 \\
  & & TRLSE  & $0.391_{.022}$ & $27.4_{2.4}$ & $68.6_{2.1}$ & & $0.035_{.008}\,(0.1)$ & $0.001_{.003}$ & $0.943_{.003}$ & $26.8_{1.2}$ & 650 \\
  & & ATLAS  & $\mathbf{0.454}_{.016}$ & $\mathbf{61.9}_{4.4}$ & $36.2_{3.2}$ & & $\mathbf{0.077}_{.009}\,(18.0)$ & $\mathbf{0.093}_{.018}$ & $0.784_{.007}$ & $\mathbf{52.0}_{1.1}$ & 538 \\
  \cmidrule(lr){2-12}

  & \multirow{3}{*}{100D}
    & AP2L   & $0.002_{.002}$ & $0.1_{0.1}$ & $7.0_{1.6}$ & & $0.010_{.003}\,(0.0)$ & $0.000_{.000}$ & $\mathbf{1.001}_{.004}$ & $3.5_{2.6}$ & 650 \\
  & & TRLSE  & $0.182_{.031}$ & $10.3_{2.0}$ & $\mathbf{79.0}_{2.4}$ & & $0.021_{.003}\,(0.1)$ & $0.000_{.000}$ & $0.928_{.002}$ & $39.2_{1.2}$ & 650 \\
  & & ATLAS  & $\mathbf{0.307}_{.038}$ & $\mathbf{20.8}_{3.6}$ & $60.6_{3.7}$ & & $\mathbf{0.051}_{.006}\,(20.3)$ & $\mathbf{0.006}_{.007}$ & $0.837_{.007}$ & $\mathbf{52.0}_{2.2}$ & 538 \\
\midrule

\multirow{9}{*}{\textbf{Rast.}} 
  & \multirow{3}{*}{10D}
    & AP2L   & $0.259_{.192}$ & $31.5_{30.5}$ & $53.3_{23.7}$ & & $0.474_{.110}\,(0.1)$ & $0.643_{.061}$ & $0.900_{.081}$ & $12.9_{11.4}$ & 650 \\
  & & TRLSE  & $0.366_{.034}$ & $26.0_{3.5}$ & $\mathbf{62.9}_{1.5}$ & & $0.489_{.051}\,(0.0)$ & $0.608_{.020}$ & $\mathbf{0.989}_{.012}$ & $30.0_{2.0}$ & 650 \\
  & & ATLAS  & $\mathbf{0.432}_{.025}$ & $\mathbf{51.1}_{10.0}$ & $39.3_{5.7}$ & & $\mathbf{0.644}_{.061}\,(0.2)$ & $\mathbf{0.729}_{.014}$ & $0.802_{.031}$ & $\mathbf{48.7}_{3.0}$ & 538 \\
  \cmidrule(lr){2-12}

  & \multirow{3}{*}{50D}
    & AP2L   & $0.245_{.201}$ & $21.1_{21.1}$ & $61.4_{14.0}$ & & $0.221_{.049}\,(0.0)$ & $0.388_{.067}$ & $\mathbf{0.991}_{.015}$ & $12.8_{11.7}$ & 650 \\
  & & TRLSE  & $0.396_{.019}$ & $27.1_{1.8}$ & $\mathbf{73.7}_{1.3}$ & & $0.257_{.032}\,(0.0)$ & $0.443_{.015}$ & $0.952_{.004}$ & $35.2_{1.5}$ & 650 \\
  & & ATLAS  & $\mathbf{0.471}_{.024}$ & $\mathbf{38.8}_{4.6}$ & $60.9_{3.3}$ & & $\mathbf{0.339}_{.029}\,(0.9)$ & $\mathbf{0.541}_{.021}$ & $0.989_{.021}$ & $\mathbf{42.4}_{2.6}$ & 538 \\
  \cmidrule(lr){2-12}

  & \multirow{3}{*}{100D}
    & AP2L   & $0.227_{.144}$ & $14.9_{9.6}$ & $\mathbf{67.1}_{2.7}$ & & $0.150_{.029}\,(0.0)$ & $0.276_{.054}$ & $0.995_{.005}$ & $15.5_{10.2}$ & 650 \\
  & & TRLSE  & $0.000_{.000}$ & $0.0_{0.0}$ & -- & & $0.183_{.021}\,(0.0)$ & $0.327_{.015}$ & $0.922_{.002}$ & $47.1_{3.1}$ & 650 \\
  & & ATLAS  & $\mathbf{0.273}_{.036}$ & $\mathbf{18.7}_{3.8}$ & $52.9_{3.8}$ & & $\mathbf{0.304}_{.023}\,(3.6)$ & $\mathbf{0.515}_{.017}$ & $\mathbf{1.032}_{.018}$ & $\mathbf{50.0}_{1.7}$ & 538 \\
\midrule

\multirow{9}{*}{\textbf{Rosen.}} 
  & \multirow{3}{*}{10D}
    & AP2L   & $0.019_{.044}$ & $10.5_{30.4}$ & $50.3_{49.7}$ & & $0.676_{.127}\,(0.0)$ & $0.742_{.168}$ & $0.993_{.030}$ & $4.0_{3.1}$ & 650 \\
  & & TRLSE  & $0.682_{.016}$ & $57.6_{2.3}$ & $\mathbf{83.8}_{2.0}$ & & $0.812_{.059}\,(0.0)$ & $0.811_{.015}$ & $\mathbf{1.086}_{.015}$ & $32.8_{5.8}$ & 650 \\
  & & ATLAS  & $\mathbf{0.700}_{.015}$ & $\mathbf{79.4}_{4.1}$ & $63.1_{4.7}$ & & $\mathbf{0.879}_{.045}\,(0.0)$ & $\mathbf{0.913}_{.014}$ & $0.959_{.020}$ & $\mathbf{33.1}_{5.9}$ & 538 \\
  \cmidrule(lr){2-12}

  & \multirow{3}{*}{50D}
    & AP2L   & $0.006_{.024}$ & $3.6_{18.3}$ & $60.7_{45.8}$ & & $0.326_{.058}\,(0.0)$ & $0.528_{.039}$ & $\mathbf{0.997}_{.007}$ & $3.8_{2.8}$ & 650 \\
  & & TRLSE  & $0.185_{.024}$ & $10.6_{1.5}$ & $75.2_{2.2}$ & & $0.536_{.053}\,(0.0)$ & $0.712_{.015}$ & $0.921_{.004}$ & $37.2_{2.2}$ & 650 \\
  & & ATLAS  & $\mathbf{0.300}_{.055}$ & $\mathbf{18.8}_{4.6}$ & $\mathbf{78.4}_{4.7}$ & & $\mathbf{0.730}_{.027}\,(3.0)$ & $\mathbf{0.852}_{.008}$ & $0.846_{.006}$ & $\mathbf{59.5}_{1.1}$ & 538 \\
  \cmidrule(lr){2-12}

  & \multirow{3}{*}{100D}
    & AP2L   & $0.006_{.022}$ & $3.6_{18.3}$ & $4.1_{3.8}$ & & $0.250_{.055}\,(0.0)$ & $0.417_{.065}$ & $\mathbf{0.998}_{.008}$ & $3.5_{2.6}$ & 650 \\
  & & TRLSE  & $0.000_{.000}$ & $0.0_{0.0}$ & -- & & $0.397_{.025}\,(0.0)$ & $0.603_{.011}$ & $0.937_{.001}$ & $\mathbf{66.3}_{1.7}$ & 650 \\
  & & ATLAS  & $\mathbf{0.461}_{.028}$ & $\mathbf{41.0}_{5.7}$ & $\mathbf{55.0}_{8.4}$ & & $\mathbf{0.844}_{.028}\,(32.8)$ & $\mathbf{0.914}_{.011}$ & $0.839_{.007}$ & $47.6_{2.8}$ & 538 \\
\bottomrule
\end{tabular}
\begin{tablenotes} 
\item[1] Sampling deep inside the super level set sacrifices average dispersion, since samples there are closer together than samples along the boundary. As ATLAS goes deep into the super level set to produce a set with more representative margin spread, its set’s coverage decreases on average, which is why our dispersion spread is somewhat lower than our baselines. Although it loses coverage, it gains coverage of adversarial scores that would not otherwise be recoverable with the other baselines or uniform sampling. At 50D and 100D, ATLAS recovers adversarial scores that are higher than every sample in the super level test set.
\end{tablenotes}
\end{threeparttable}
\end{table}

\section{Image Experiments: Mapping Adversarial Regions through Latent-Space Search}
\label{sec:image_attack_results}
We evaluate whether ATLAS can recover adversarial sets for image-classification models. Following the generator-constrained attack setting of \citep{song2018constructing}, we search the latent space of a fixed class-conditional generator. We consider only the space of natural images for candidate adversaries and learn a value function, $\valuefunc(\inputsample)$ whose super level set represents latent samples $\latentinputsample \in \latentSpace$ of a fixed generator, $\generator(\latentinputsample, \classcondition) \rightarrow \generatoroutput$ in which the decoded image represents a chosen source class $\classcondition \in \mathbf{R}^{\numMNISTclasses}$ that gets classified by the target as a chosen target class. For each dataset--model setting, we evaluate 20 source--target class pairs. We restrict the latent domain to a bounded region around the generator prior mean, $\latentSpace = \left\{ \latentinputsample \in \mathbb{R}^d \;:\; \Vert{}\latentinputsample - \priormean\Vert{}_\infty \le \zscore\priorsigma \right\}$, where $\zscore$ indicated $\zscore$-deviations away from the latent prior's mean $\priormean$. To align successful targeted attacks with a super-zero level set, we define the negated objective function which takes from Carlini\&Wagner \citep{carlini2017towards}:
\begin{equation}
\begin{aligned}
\adversarialobjective(\inputsample'; t) = \logitfunc_t(\inputsample') - \max_{i \neq t} \logitfunc_i(\inputsample')
\end{aligned}
\label{eq:logits-margin}
\end{equation}
where $\logitfunc(\inputsample) = \logitfuncoutput$ be the output of all neural layers of the target system (the logits before softmax is applied) so $\targetsystem(\inputsample) = \text{softmax}(\logitfunc(\inputsample))$. Under this formulation, the successful targeted attack region $\genericadversarialset \coloneqq \{\inputsample \in \inputspace: \advfunclandscape(\inputsample) \geq \threshold\}$ corresponds directly to the non-negative super-level set of Eq. \ref{eq:logits-margin}. We further use a nearest neighbor metric \citep{clark1954distance} to measure the ratio of mean nearest-neighbour distance to mean pairwise distance within the set:
\begin{equation}
\begin{aligned}
\clump(\genericadversarialset) = \frac{\frac{1}{M} \sum_{i}\min_{j\neq i}\|\genericadversarialset_i - \genericadversarialset_j\|_2}{\dispersionmetric(\genericadversarialset)}
\end{aligned}
\label{eq:clump}
\end{equation}
Values near 1 indicate points spread evenly through the region they occupy; values near 0 indicate the set collapses into a few tight clusters. We compare ATLAS to NES, SignHunter, BayesOpt, and uniform random sampling (see Appendix \ref{sec:Image Appendix} for details) and report the results in Table~\ref{tab:atlas_audit_summary}. 

At 650-budget, the results show that ATLAS recovers the most representative adversarial set. Although ATLAS recovers fewer raw points than a couple of the baselines, it distributes its points far better than any of the other baselines. On the one hand, uniform random sampling clusters the least (by construction) with $\clump=$ 0.88-0.93 but finds the fewest adversarial points (1-19). On the other, NES and SignHunter finds the most adversarial points, but with either less spread $\dispersionmetricnorm=$ 1.14-2.43 (v. 1.51-3.82) or more clustering $\clump=$ 0.24-0.31 (v. 0.55–0.65) respectively, implying that their points collapse into smaller, tighter neighborhoods. ATLAS is the only method that achieves best on Margin J in 4 of 5 targets with good size, ASR, spatial spread and spacing. Although, it may be slightly slower to the first success than BayesOpt ($Q_{first}=$ 83–229 vs. 36–102 for BayesOpt), it spends its budget effectively finding more adversarial regions to find as many diverse model failures as possible as opposed to cheaply sampling the same attacks near already known regions. The former is much better for a robust security evaluation. Figure \ref{fig:compressed-atlas-examples} shows ATLAS's representative sets for each target and Appendix \ref{sec:security_audit} reports an example security audit.

\begin{table}[t]
\centering
\caption{\textbf{ATLAS matches the best attack success rate while returning the most representative adversarial set.} 
Each method is given $B = 650$ oracle queries per experiment, averaged over 20 source–target class pairs per target model; subscripts are standard deviations across pairs and the best value per metric and target is in bold. ASR is the fraction of pairs with at least one success; $Q_{first}$ is queries to first success, computed over succeeding pairs only. Margin J, $\dispersionmetricnorm$ and $\clump$ describe how the recovered points are distributed (higher = more of the region covered, covered more evenly). ATLAS is the only method achieving the best distribution of adversarial scores (Margin J) with good spatial spread and competitive set size. Standard CIFAR-10 and ImageNet targets are in Table~\ref{tab:image-latent-main-appendix}.}
\label{tab:atlas_audit_summary}
\begin{threeparttable}
\scriptsize
\setlength{\tabcolsep}{1.5pt}
\begin{tabular}{ll l cc c cccc}
\toprule
& & & \multicolumn{2}{c}{\textbf{Attack}} & & \multicolumn{4}{c}{\textbf{Adversarial Set}} \\
\cmidrule(lr){4-5} \cmidrule(lr){7-10}
\textbf{Data} & \textbf{Target} & \textbf{Method} & \textbf{ASR\% $\uparrow$} & \textbf{$Q_{first}$ $\downarrow$} & & \textbf{Margin J $\uparrow$} & \textbf{$\dispersionmetricnorm$ $\uparrow$} & \textbf{Adv. Count $\uparrow$} & \textbf{$\clump$ $\uparrow$} \\
\midrule

  \multirow{5}{*}{\textbf{MNIST}} & \multirow{5}{*}{Standard} & \textbf{ATLAS}         & $95$ & $164_{85}$ & & $\mathbf{0.859}_{0.091}$ & $1.94_{0.15}$ & $174.4_{109.1}$ & $0.63_{0.04}$ \\
   &  & NES                    & $75$ & $181_{162}$ & & $0.745_{0.238}$ & $1.14_{0.12}$ & $225.9_{199.2}$ & $0.58_{0.13}$ \\
   &  & SignHunter             & $\mathbf{100}$ & $60_{48}$ & & $0.816_{0.075}$ & $\mathbf{2.99}_{0.31}$ & $\mathbf{255.4}_{99.4}$ & $0.27_{0.03}$ \\
   &  & Uniform                & $26$ & $408_{211}$ & & $0.136_{0.207}$ & $2.25_{0.03}$ & $1.2_{3.5}$ & $\mathbf{0.92}_{0.10}$ \\
   &  & BayesOpt\tnote{a}      & $\mathbf{100}$ & $\mathbf{36}_{90}$ & & -- & -- & -- & -- \\
  \midrule
  \multirow{5}{*}{\textbf{CIFAR-10}} & \multirow{5}{*}{Adv.} & \textbf{ATLAS}         & $\mathbf{100}$ & $83_{81}$ & & $\mathbf{0.823}_{0.114}$ & $1.88_{0.18}$ & $203.4_{96.9}$ & $0.57_{0.03}$ \\
   &  & NES                    & $95$ & $121_{95}$ & & $0.803_{0.162}$ & $1.24_{0.13}$ & $282.6_{172.2}$ & $0.52_{0.08}$ \\
   &  & SignHunter             & $\mathbf{100}$ & $47_{67}$ & & $0.770_{0.109}$ & $\mathbf{3.00}_{0.31}$ & $\mathbf{306.6}_{142.4}$ & $0.24_{0.05}$ \\
   &  & Uniform                & $85$ & $96_{163}$ & & $0.356_{0.204}$ & $2.24_{0.09}$ & $10.9_{15.8}$ & $\mathbf{0.87}_{0.07}$ \\
   &  & BayesOpt\tnote{a}      & $\mathbf{100}$ & $\mathbf{44}_{45}$ & & -- & -- & -- & -- \\
  \midrule
  \multirow{5}{*}{\textbf{ImageNet}} & \multirow{5}{*}{Adv.} & \textbf{ATLAS}         & $\mathbf{100}$ & $162_{158}$ & & $\mathbf{0.745}_{0.206}$ & $3.82_{0.75}$ & $115.1_{104.4}$ & $0.55_{0.14}$ \\
   &  & NES                    & $85$ & $98_{133}$ & & $0.703_{0.194}$ & $2.43_{0.23}$ & $144.6_{180.2}$ & $0.61_{0.09}$ \\
   &  & SignHunter             & $85$ & $147_{150}$ & & $0.583_{0.189}$ & $3.37_{0.97}$ & $\mathbf{195.4}_{112.1}$ & $0.26_{0.08}$ \\
   &  & Uniform                & $55$ & $\mathbf{25}_{118}$ & & $0.391_{0.264}$ & $\mathbf{4.48}_{0.12}$ & $15.3_{39.9}$ & $\mathbf{0.93}_{0.05}$ \\
   &  & BayesOpt\tnote{a}      & $\mathbf{100}$ & $102_{121}$ & & -- & -- & -- & -- \\

\bottomrule
\end{tabular}
\begin{tablenotes}
\item[a] BayesOpt is run in its published configuration, which stops at its first
adversarial point, so it is read for ASR and $Q_{first}$ only: its Count is $1$ by
construction.
\end{tablenotes}
\end{threeparttable}
\end{table}


\begin{figure}[!t]
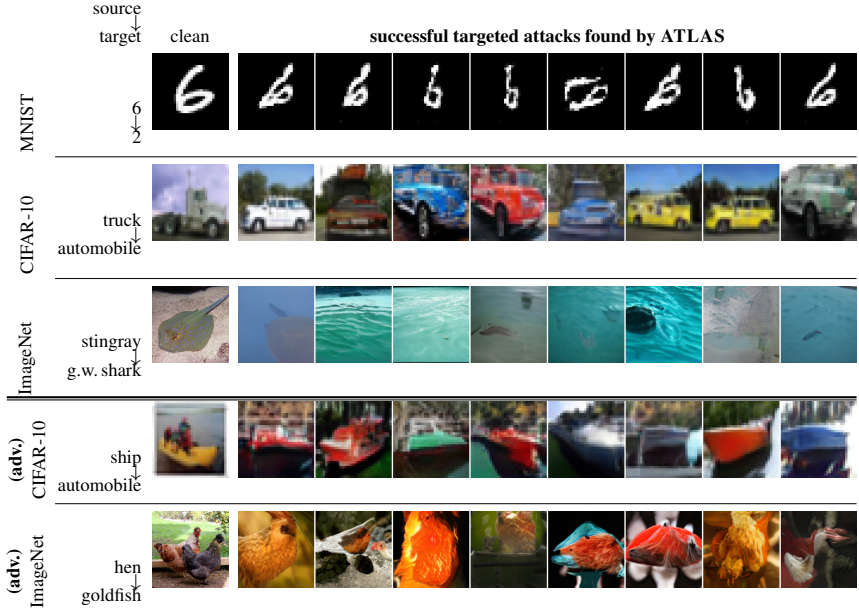

  \centering
  {
  \setlength{\tabcolsep}{2.0pt}
  \renewcommand{\arraystretch}{0.92}
  \renewcommand{\dslabel}[1]{%
      \rotatebox[origin=c]{90}{\scriptsize\shortstack{#1}}%
    }

  \begin{tabular}{@{}c r c c@{}}

    &
    \scriptsize
    \begin{tabular}[b]{@{}r@{}}
      source\\[-2pt]
      $\downarrow$\\[-2pt]
      target
    \end{tabular}
    &
    \scriptsize clean
    &
    \scriptsize\textbf{successful targeted attacks found by \textsc{ATLAS}}
    \tabularnewline[1pt]


    \dslabel{MNIST}
    & \pairlabel{6}{2}
    & \atlasclean{standard_mnist_b4}
    & \atlasadv{standard_mnist_b4}
    \tabularnewline



    \cmidrule(l{1pt}r{0pt}){2-4}

    \dslabel{CIFAR-10}
    & \pairlabel{truck}{automobile}
    & \atlasclean{standard_cifar10_b0}
    & \atlasadv{standard_cifar10_b0}
    \tabularnewline



    \cmidrule(l{1pt}r{0pt}){2-4}

    \dslabel{ImageNet}
    & \pairlabel{stingray}{g.w.\,shark}
    & \atlasclean{standard_imagenet_b4}
    & \atlasadv{standard_imagenet_b4}
    \tabularnewline


    \specialrule{0.8pt}{0.5pt}{0.2pt}
    \specialrule{0.4pt}{0.2pt}{0.5pt}

    \dslabel{\textbf{(adv.)}\\CIFAR-10}


    & \pairlabel{ship}{automobile}
    & \atlasclean{adv_trained_cifar10_b2}
    & \atlasadv{adv_trained_cifar10_b2}
    \tabularnewline

    \cmidrule(l{1pt}r{0pt}){2-4}

    \dslabel{\textbf{(adv.)}\\ImageNet}
    & \pairlabel{hen}{goldfish}
    & \atlasclean{adv_trained_imagenet_b2}
    & \atlasadv{adv_trained_imagenet_b2}
    \tabularnewline



  \end{tabular}
  }

  \vspace{1pt}

  \caption{\textbf{Adversarial sets expose shared failure modes that single attacks cannot.} Representative examples discovered by \textsc{ATLAS} under a 650-query budget; rows below the double rule are adversarially trained targets. Each row is one source$\rightarrow$target pair: the leftmost image is a clean source-class sample, the remainder are members of the recovered adversarial set. Read jointly, each set reveals what the target model keys on. For stingray$\rightarrow$great white (g.w.) shark, every recovered image contains an open ocean rather than a shark's physical features; for goldfish targets, the recovered images contain flower petals and feathers perhaps standing in for fin-like features. These patterns point to areas for more needed study and analysis for the target model. Appendix~\ref{sec:interpret_image_results} gives the full grid and analysis.}

  \label{fig:compressed-atlas-examples}
\end{figure}

\section{Conclusion}
\label{Sec: Conclusion}
In conclusion, this paper presents \textsc{ATLAS}, a framework that reformulates adversarial discovery as a region-recovery problem, leverages active learning and level-set estimation for this problem, and introduces new mechanisms that are specifically required for adversarial-set and adversarial classification construction. In this work, we find that ATLAS can construct more diverse adversarial sets, which provides a useful foundation for building automated red-teaming tools. There is still some variance in performance between function landscapes. Future work involves investigating on more targets and training target models on ATLAS's generated adversarial sets. 
\clearpage
{
    \small
    \bibliographystyle{plainnat}
    \bibliography{main_atlas}
}
\appendix
\section{Full Toy Experiment Evaluation}
\label{sec: appendix-ext}
This supplementary appendix contains the comprehensive, fully expanded empirical tables across all benchmark geometry configurations: Ackley, Rastrigin, and Rosenbrock across $10\text{D}$, $50\text{D}$, and $100\text{D}$ settings. Our adversarial objective function for this experiment is: $\adversarialobjective(\inputsample) = -\toyfunction(\inputsample-\randomoffset)$, where $\toyfunction$ is the function being minimized and $\randomoffset$ is a design parameter used to offset the minimizer. Each method is run with $\expbudget$ queries. Dataset sizes, calibration set sizes, and the number of trust regions are kept the same across methods. We use a 1:1 ratio of saturation to boundary regions and maintain 4 trust regions in all experiments. Full settings, including footnotes on how weights were chosen and where they deviate, are in Table~\ref{tab:hyperparams}. We do not alter the baseline implementations, since doing so would require re-tuning them. ATLAS includes several speed-ups to its optimization (Appendix~\ref{sec: Speedups}). Table~\ref{tab:calibration-ablation-appendix} reports the full calibration sweep and Table~\ref{tab:saturation-ablation-appendix} the full boundary/saturation ablation. Table~\ref{tab:default-table} repeats the main benchmark using a single default weight setting ($\boundaryweight{=}2$, $\superlevelprobweight{=}40$) shared across all benchmarks in place of per-benchmark weights. ATLAS retains the best recall in 8 of 9 settings under these defaults.

\begin{table}[h]
\begin{threeparttable}
\centering
\caption{\textbf{Experiment settings for ATLAS and baselines.} Shared settings are identical across methods; matched settings are set equal between ATLAS and the baseline that has the counterpart. Baseline-specific values follow each method's reference implementation. $d$-dependent values are listed as 10D / 50D / 100D. Baseline-specific values follow each method's reference implementation. Weights $\boundaryweight$ and $\superlevelprobweight$ are set per benchmark to match the scale of $\beta\sigma$ (footnote 2). With a single default setting ($\boundaryweight{=}2$, $\superlevelprobweight{=}40$) across all benchmarks, ATLAS still achieves the best recall in 8 of 9 and the best F1 in 7 of 9 (Table~\ref{tab:default-table}), so the reported comparison does not rest on this choice. $\dimension$-dependent values are listed as 10D / 50D / 100D.}
\label{tab:hyperparams}
\scriptsize
\setlength{\tabcolsep}{3pt}
\begin{tabular}{@{}llll@{}}
\toprule
Parameter & ATLAS & TRLSE & AP2L \\
\midrule
\multicolumn{4}{@{}l}{\textit{Experimental Params}} \\
query budget $\globalsamplingbudget$ & 650 (112 calib.\ + 538) & 650 & 650 (+112 calib.)\tnote{1} \\
seeds; test sets & \multicolumn{3}{l}{30; 50K ($\superlevelset$), 1M (Mix)} \\
super-level fraction; functions & \multicolumn{3}{l}{0.05; same domain and minimizer offset} \\
points per acquisition & 1 & 1 & 1 \\
surrogate & GP, Mat\'ern-5/2 & GP, Mat\'ern-5/2 (ARD) & GP, Mat\'ern-5/2 \\
\midrule
\multicolumn{4}{@{}l}{\textit{Common Params}} \\
trust regions & 2 bound.\ + 2 sat. & 4 & -- \\
uncertainty weight $\beta$ & 1.96\tnote{4} & 1.96\tnote{3} & 1.28\tnote{3} \\
calibration samples & 112 & -- & 112 \\
region volume $\volume$ & 0.1 / 0.1 / $0.5^{100}$ (fixed) & init 0.033/0.033/$0.5^{100}$; max 0.1/0.1/0.01 & -- \\
candidate pool\tnote{5} & 20K & 100K (Sobol) & 100K \\
\midrule
\multicolumn{4}{@{}l}{\textit{Method-specific}} \\
ATLAS & \multicolumn{3}{p{0.78\linewidth}}{$\boundaryweight{=}2$ (Rast.\ 100D: 16); $\superlevelprobweight$ per setting\tnote{2}; $\localboundarymixingweight{=}0.5$; $\penaltyfuncweight{=}10$, $w_g{=}0.05$, $\seedseparation{=}2\trustregionlength$; GP refit every 4 rounds} \\
TRLSE & \multicolumn{3}{p{0.78\linewidth}}{$V_{\min} = V_{\text{init}}/2$; regions persist and resize; classify at threshold 0} \\
AP2L & \multicolumn{3}{p{0.78\linewidth}}{$\hat{e}$ threshold 0.3; decay 0.95; conformal $\alpha{=}0.05$} \\
\bottomrule
\end{tabular}
\begin{tablenotes}
\item[1] AP2L's calibration set is drawn outside its query budget; ATLAS's is counted against it.
\item[2] For Rosenbrock and Rastrigin, $\superlevelprobweight$ scales with dimension. Ack.\ 40; Rosen.\ $10^3$ / $10^5$ / $5{\times}10^6$; Rast.\ 100 / (40) / $10^4$. The saturation acquisition (Eq.~\ref{eq:global-sat-acq}) adds $\superlevelprobweight\postprob$ to $\globaluncertaintyparam\localuncertainty(\inputsample)$, where $\localuncertainty$ is expressed in the target function's output units. These units differ by roughly five orders of magnitude across our benchmarks and grow with dimension, so no single $\superlevelprobweight$ balances the two terms across settings. Therefore, to avoid a saturation acquisition dominated by the exploration term $\globaluncertaintyparam\localuncertainty(\inputsample)$, we set $\superlevelprobweight$ per benchmark to be at least the observed magnitude of $\globaluncertaintyparam\localuncertainty$ ($\superlevelprobweight/\globaluncertaintyparam\localuncertainty \in [1, 140]$), yielding the values above. Rastrigin 50D departs from this ($\superlevelprobweight = 40$) because raising $\superlevelprobweight$ to that scale shifted budget from boundary sampling and reduced F1 (0.54 to 0.20). With a single default setting ($\boundaryweight{=}2$, $\superlevelprobweight{=}40$) for all benchmarks (Table~\ref{tab:default-table}), ATLAS still leads recall in 8 of 9 settings and F1 in 7 of 9, so the reported comparison does not rest on this choice.
\item[3] Method's original setting.
\item[4] Matched to TRLSE's setting.
\item[5] ATLAS draws from a smaller candidate pool (20K vs. 100K), which if anything disadvantages it
\end{tablenotes}
\end{threeparttable}
\end{table}

\begin{table}[htbp]
\centering
\caption{\textbf{In general, calibration lets ATLAS learn a better model with fewer samples.} We vary the number of calibration samples $N_C$ drawn from the fixed total budget $B = 650$, leaving $B - N_C$ for sampling. We evaluate over 30 seeds. A small calibration set ($N_C$ = 50--100, 8--15\% of B) gives the best or near-best F1 in every experiment and higher recall than no calibration in all nine. Thus, for our main experiment, we use $N_C = 112 (17\%)$, inside of this range. The benefit grows with dimension. At 10D, the GP may learn well without calibration (Rosenbrock 10D), while at 100D, removing calibration may collapse performance (F1 for Rastrigin 0.003, Rosenbrock 0.100).  While calibration is helpful, over-allocating the budget to calibration (58\%) degrades performance.}
\label{tab:calibration-ablation-appendix}
\begin{threeparttable}
\scriptsize
\setlength{\tabcolsep}{5pt}
\begin{tabular}{lll ccc}
\toprule
\textbf{Func} & \textbf{Dim} & \textbf{Calib. $N_C$ (\% of B)} & \textbf{F1 (Mix) $\uparrow$} & \textbf{Rec $\superlevelset$ \% $\uparrow$} & \textbf{Prec (Mix) \% $\uparrow$}\tnote{1} \\
\midrule

\multirow{15}{*}{\textbf{Ackley}} 
  & \multirow{5}{*}{10D}
    & $0$ (0\%) & $0.678_{.012}$ & $60.3_{2.1}$ & $\mathbf{77.5}_{1.9}$ \\
  & & $50$ (8\%) & $0.663_{.022}$ & $62.7_{5.4}$ & $71.0_{4.1}$ \\
  & & $100$ (15\%) & $\mathbf{0.687}_{.025}$ & $\mathbf{66.6}_{5.0}$ & $71.4_{3.8}$ \\
  & & $250$ (38\%) & $0.664_{.016}$ & $62.2_{5.6}$ & $72.1_{5.8}$ \\
  & & $375$ (58\%) & $0.662_{.020}$ & $63.3_{5.3}$ & $70.3_{5.4}$ \\
  \cmidrule(lr){2-6}
  & \multirow{5}{*}{50D}
    & $0$ (0\%) & $0.226_{.075}$ & $14.1_{6.4}$ & $\mathbf{72.0}_{9.1}$ \\
  & & $50$ (8\%) & $0.448_{.019}$ & $\mathbf{63.5}_{4.6}$ & $34.9_{3.3}$ \\
  & & $100$ (15\%) & $0.450_{.018}$ & $63.1_{4.4}$ & $35.2_{3.3}$ \\
  & & $250$ (38\%) & $\mathbf{0.460}_{.012}$ & $59.8_{5.2}$ & $37.8_{3.4}$ \\
  & & $375$ (58\%) & $0.367_{.075}$ & $32.2_{10.8}$ & $47.2_{5.0}$ \\
  \cmidrule(lr){2-6}
  & \multirow{5}{*}{100D}
    & $0$ (0\%) & $0.163_{.020}$ & $9.2_{1.3}$ & $\mathbf{77.1}_{2.7}$ \\
  & & $50$ (8\%) & $\mathbf{0.322}_{.069}$ & $\mathbf{22.8}_{6.2}$ & $60.3_{6.3}$ \\
  & & $100$ (15\%) & $0.305_{.056}$ & $21.0_{5.7}$ & $60.6_{5.1}$ \\
  & & $250$ (38\%) & $0.199_{.050}$ & $12.1_{3.7}$ & $63.4_{5.4}$ \\
  & & $375$ (58\%) & $0.029_{.028}$ & $1.5_{1.5}$ & $72.9_{10.6}$ \\
\midrule

\multirow{15}{*}{\textbf{Rast.}} 
  & \multirow{5}{*}{10D}
    & $0$ (0\%) & $0.390_{.020}$ & $30.0_{2.5}$ & $56.2_{3.4}$ \\
  & & $50$ (8\%) & $\mathbf{0.430}_{.023}$ & $\mathbf{50.6}_{12.4}$ & $41.2_{10.3}$ \\
  & & $100$ (15\%) & $0.421_{.050}$ & $35.1_{11.1}$ & $\mathbf{61.0}_{10.9}$ \\
  & & $250$ (38\%) & $0.428_{.018}$ & $49.8_{9.4}$ & $40.1_{9.1}$ \\
  & & $375$ (58\%) & $0.422_{.021}$ & $48.6_{6.0}$ & $37.9_{3.6}$ \\
  \cmidrule(lr){2-6}
  & \multirow{5}{*}{50D}
    & $0$ (0\%) & $0.304_{.022}$ & $19.2_{1.8}$ & $73.8_{1.9}$ \\
  & & $50$ (8\%) & $0.471_{.039}$ & $\mathbf{39.8}_{7.7}$ & $60.4_{5.6}$ \\
  & & $100$ (15\%) & $\mathbf{0.473}_{.018}$ & $39.0_{3.2}$ & $60.8_{2.5}$ \\
  & & $250$ (38\%) & $0.420_{.078}$ & $32.1_{8.6}$ & $65.9_{6.4}$ \\
  & & $375$ (58\%) & $0.191_{.140}$ & $12.1_{9.8}$ & $\mathbf{82.8}_{10.1}$ \\
  \cmidrule(lr){2-6}
  & \multirow{5}{*}{100D}
    & $0$ (0\%) & $0.003_{.002}$ & $0.2_{0.1}$ & $\mathbf{94.6}_{3.2}$ \\
  & & $50$ (8\%) & $\mathbf{0.288}_{.051}$ & $\mathbf{20.3}_{5.2}$ & $54.0_{6.1}$ \\
  & & $100$ (15\%) & $0.278_{.035}$ & $19.1_{3.5}$ & $53.7_{4.3}$ \\
  & & $250$ (38\%) & $0.251_{.030}$ & $16.8_{2.7}$ & $51.0_{3.3}$ \\
  & & $375$ (58\%) & $0.214_{.046}$ & $13.7_{3.8}$ & $51.6_{4.1}$ \\
\midrule

\multirow{15}{*}{\textbf{Rosen.}} 
  & \multirow{5}{*}{10D}
    & $0$ (0\%) & $\mathbf{0.732}_{.006}$ & $73.1_{2.5}$ & $73.4_{2.1}$ \\
  & & $50$ (8\%) & $0.701_{.020}$ & $\mathbf{80.9}_{4.7}$ & $62.3_{5.6}$ \\
  & & $100$ (15\%) & $0.698_{.015}$ & $79.3_{3.9}$ & $62.6_{4.2}$ \\
  & & $250$ (38\%) & $0.691_{.013}$ & $68.5_{5.4}$ & $70.3_{4.9}$ \\
  & & $375$ (58\%) & $0.643_{.033}$ & $56.7_{8.1}$ & $\mathbf{76.5}_{6.3}$ \\
  \cmidrule(lr){2-6}
  & \multirow{5}{*}{50D}
    & $0$ (0\%) & $0.333_{.038}$ & $21.6_{3.7}$ & $75.4_{3.3}$ \\
  & & $50$ (8\%) & $\mathbf{0.342}_{.065}$ & $\mathbf{23.1}_{7.1}$ & $73.7_{7.2}$ \\
  & & $100$ (15\%) & $0.312_{.057}$ & $20.0_{5.2}$ & $76.8_{5.6}$ \\
  & & $250$ (38\%) & $0.268_{.042}$ & $16.2_{3.1}$ & $82.6_{3.5}$ \\
  & & $375$ (58\%) & $0.209_{.046}$ & $12.0_{3.2}$ & $\mathbf{87.0}_{4.0}$ \\
  \cmidrule(lr){2-6}
  & \multirow{5}{*}{100D}
    & $0$ (0\%) & $0.100_{.035}$ & $5.3_{2.0}$ & $\mathbf{90.2}_{1.6}$ \\
  & & $50$ (8\%) & $\mathbf{0.457}_{.037}$ & $\mathbf{43.6}_{9.6}$ & $52.2_{9.9}$ \\
  & & $100$ (15\%) & $0.455_{.041}$ & $41.9_{8.0}$ & $52.8_{8.9}$ \\
  & & $250$ (38\%) & $0.447_{.026}$ & $39.0_{4.6}$ & $53.9_{7.1}$ \\
  & & $375$ (58\%) & $0.423_{.025}$ & $36.1_{4.1}$ & $52.7_{7.5}$ \\
\bottomrule
\end{tabular}
\begin{tablenotes}
\item[1] No calibration attains high precision by predicting few super-level points, at the cost of recall. This is undesirable for a security analysis tool, whose purpose is to catch as many adversarial inputs as possible.
\end{tablenotes}
\end{threeparttable}
\end{table}

\clearpage

\begin{table}[t]
\begin{threeparttable}
\centering
\caption{\textbf{Saturation and boundary sampling allow ATLAS to produce a more diverse adversarial set while having robust model estimation across functions and dimensions.} All variants use same sampling budget (B=538) and 4 trust regions. ATLAS-b uses only boundary regions, ATLAS-s only saturation, and ATLAS a 1:1 split. ATLAS-s and ATLAS have the best margin spread and depth. Saturation, as designed, helps harvest adversarial examples, winning nSuper/B in 7 of 9 cells, often at about 90\%. ATLAS-b has the best or near best spatial spread and precision\tnote{1} in most experiments, helping us explore widely and find boundary points, but loses at margin spread and depth. ATLAS is the only variant that is robust across all landscapes with model estimation and adversarial set quality. This highlights the need to balance both sampling designs.}
\label{tab:saturation-ablation-appendix}
\scriptsize
\setlength{\tabcolsep}{2pt}
\begin{tabular}{lll ccc c ccccc}
\toprule
& & & \multicolumn{3}{c}{\textbf{Learned Model Metrics}} & & \multicolumn{5}{c}{\textbf{Sampling Set Metrics}} \\
\cmidrule(lr){4-6} \cmidrule(lr){8-12}
\textbf{Func} & \textbf{Dim} & \textbf{Arm} & \textbf{F1 (Mix) $\uparrow$} & \textbf{Rec $\superlevelset$ \% $\uparrow$} & \textbf{Prec (Mix) \% $\uparrow$} & & \textbf{mV ($\%$) $\uparrow$} & \textbf{Margin J $\uparrow$} & \textbf{$\dispersionmetricnorm$ $\uparrow$} & \textbf{nSuper/B (\%) $\uparrow$} & \textbf{B} \\
\midrule

\multirow{9}{*}{\textbf{Ackley}} 
  & \multirow{3}{*}{10D}
    & \textsf{ATLAS-b} & $0.613_{.060}$ & $\mathbf{79.6}_{4.6}$ & $51.1_{10.3}$ & & $0.126_{.043}\,(0.0)$ & $0.074_{.016}$ & $\mathbf{0.973}_{.008}$ & $38.8_{2.0}$ & 538 \\
  & & \textsf{ATLAS-s} & $0.466_{.047}$ & $71.3_{8.8}$ & $35.4_{6.7}$ & & $\mathbf{0.315}_{.045}\,(0.0)$ & $0.358_{.011}$ & $0.734_{.018}$ & $\mathbf{97.6}_{0.4}$ & 538 \\
  & & \textsf{ATLAS}   & $\mathbf{0.664}_{.020}$ & $63.6_{4.9}$ & $\mathbf{69.8}_{4.1}$ & & $0.304_{.048}\,(0.0)$ & $\mathbf{0.368}_{.016}$ & $0.735_{.033}$ & $53.7_{3.3}$ & 538 \\
  \cmidrule(lr){2-12}
  & \multirow{3}{*}{50D}
    & \textsf{ATLAS-b} & $0.010_{.054}$ & $0.6_{3.4}$ & -- & & $0.038_{.009}\,(0.1)$ & $0.001_{.003}$ & $\mathbf{0.929}_{.006}$ & $50.9_{2.4}$ & 538 \\
  & & \textsf{ATLAS-s} & $\mathbf{0.466}_{.012}$ & $\mathbf{65.0}_{3.7}$ & $\mathbf{36.5}_{2.5}$ & & $\mathbf{0.080}_{.010}\,(19.1)$ & $\mathbf{0.097}_{.019}$ & $0.793_{.010}$ & $\mathbf{90.7}_{2.8}$ & 538 \\
  & & \textsf{ATLAS}   & $0.454_{.016}$ & $61.9_{4.4}$ & $36.2_{3.2}$ & & $0.077_{.009}\,(18.0)$ & $0.093_{.018}$ & $0.784_{.007}$ & $52.0_{1.1}$ & 538 \\
  \cmidrule(lr){2-12}
  & \multirow{3}{*}{100D}
    & \textsf{ATLAS-b} & $0.223_{.082}$ & $14.0_{6.0}$ & $\mathbf{68.9}_{9.1}$ & & $0.039_{.022}\,(1.9)$ & $0.001_{.004}$ & $\mathbf{0.913}_{.026}$ & $48.3_{6.5}$ & 538 \\
  & & \textsf{ATLAS-s} & $\mathbf{0.311}_{.067}$ & $\mathbf{21.6}_{6.4}$ & $60.4_{4.8}$ & & $\mathbf{0.053}_{.005}\,(21.9)$ & $\mathbf{0.006}_{.005}$ & $0.825_{.010}$ & $\mathbf{85.7}_{2.8}$ & 538 \\
  & & \textsf{ATLAS}   & $0.307_{.038}$ & $20.8_{3.6}$ & $60.6_{3.7}$ & & $0.051_{.006}\,(20.3)$ & $\mathbf{0.006}_{.007}$ & $0.837_{.007}$ & $52.0_{2.2}$ & 538 \\
\midrule

\multirow{9}{*}{\textbf{Rast.}} 
  & \multirow{3}{*}{10D}
    & \textsf{ATLAS-b} & $0.409_{.096}$ & $46.8_{16.6}$ & $\mathbf{43.9}_{13.1}$ & & $0.456_{.053}\,(0.0)$ & $0.630_{.016}$ & $\mathbf{0.976}_{.026}$ & $37.2_{5.4}$ & 538 \\
  & & \textsf{ATLAS-s} & $0.414_{.027}$ & $41.7_{7.8}$ & $43.5_{8.3}$ & & $\mathbf{0.655}_{.047}\,(0.1)$ & $\mathbf{0.765}_{.012}$ & $0.669_{.032}$ & $\mathbf{63.9}_{4.5}$ & 538 \\
  & & \textsf{ATLAS}   & $\mathbf{0.432}_{.025}$ & $\mathbf{51.1}_{10.0}$ & $39.3_{5.7}$ & & $0.644_{.061}\,(0.2)$ & $0.729_{.014}$ & $0.802_{.031}$ & $48.7_{3.0}$ & 538 \\
  \cmidrule(lr){2-12}
  & \multirow{3}{*}{50D}
    & \textsf{ATLAS-b} & $0.298_{.087}$ & $19.0_{7.1}$ & $\mathbf{79.8}_{6.2}$ & & $0.308_{.024}\,(0.3)$ & $0.512_{.030}$ & $0.977_{.029}$ & $\mathbf{51.5}_{4.6}$ & 538 \\
  & & \textsf{ATLAS-s} & $0.414_{.046}$ & $29.7_{4.8}$ & $71.0_{4.1}$ & & $0.312_{.048}\,(0.6)$ & $0.515_{.021}$ & $\mathbf{1.019}_{.015}$ & $33.1_{3.0}$ & 538 \\
  & & \textsf{ATLAS}   & $\mathbf{0.471}_{.024}$ & $\mathbf{38.8}_{4.6}$ & $60.9_{3.3}$ & & $\mathbf{0.339}_{.029}\,(0.9)$ & $\mathbf{0.541}_{.021}$ & $0.989_{.021}$ & $42.4_{2.6}$ & 538 \\
  \cmidrule(lr){2-12}
  & \multirow{3}{*}{100D}
    & \textsf{ATLAS-b} & $0.255_{.059}$ & $16.2_{5.1}$ & $67.1_{5.7}$ & & $0.119_{.063}\,(0.0)$ & $0.249_{.128}$ & $0.816_{.143}$ & $1.5_{1.1}$ & 538 \\
  & & \textsf{ATLAS-s} & $0.011_{.010}$ & $0.6_{0.5}$ & $\mathbf{83.2}_{8.1}$ & & $\mathbf{0.317}_{.017}\,(4.2)$ & $\mathbf{0.532}_{.013}$ & $\mathbf{1.080}_{.021}$ & $\mathbf{90.9}_{2.0}$ & 538 \\
  & & \textsf{ATLAS}   & $\mathbf{0.273}_{.036}$ & $\mathbf{18.7}_{3.8}$ & $52.9_{3.8}$ & & $0.304_{.023}\,(3.6)$ & $0.515_{.017}$ & $1.032_{.018}$ & $50.0_{1.7}$ & 538 \\
\midrule

\multirow{9}{*}{\textbf{Rosen.}} 
  & \multirow{3}{*}{10D}
    & \textsf{ATLAS-b} & $0.650_{.057}$ & $53.7_{10.4}$ & $\mathbf{85.7}_{6.5}$ & & $0.817_{.050}\,(0.0)$ & $0.858_{.015}$ & $\mathbf{1.041}_{.014}$ & $\mathbf{49.4}_{2.8}$ & 538 \\
  & & \textsf{ATLAS-s} & $0.691_{.011}$ & $76.0_{4.7}$ & $63.8_{4.0}$ & & $0.871_{.050}\,(0.1)$ & $0.885_{.026}$ & $0.852_{.034}$ & $25.7_{9.3}$ & 538 \\
  & & \textsf{ATLAS}   & $\mathbf{0.700}_{.015}$ & $\mathbf{79.4}_{4.1}$ & $63.1_{4.7}$ & & $\mathbf{0.879}_{.045}\,(0.0)$ & $\mathbf{0.913}_{.014}$ & $0.959_{.020}$ & $33.1_{5.9}$ & 538 \\
  \cmidrule(lr){2-12}
  & \multirow{3}{*}{50D}
    & \textsf{ATLAS-b} & $0.165_{.108}$ & $9.7_{6.7}$ & $84.4_{6.3}$ & & $0.529_{.080}\,(0.0)$ & $0.650_{.019}$ & $\mathbf{0.958}_{.009}$ & $38.8_{3.5}$ & 538 \\
  & & \textsf{ATLAS-s} & $0.096_{.055}$ & $5.2_{3.2}$ & $\mathbf{90.2}_{4.6}$ & & $0.716_{.031}\,(1.1)$ & $0.794_{.015}$ & $0.822_{.007}$ & $\mathbf{91.3}_{3.2}$ & 538 \\
  & & \textsf{ATLAS}   & $\mathbf{0.300}_{.055}$ & $\mathbf{18.8}_{4.6}$ & $78.4_{4.7}$ & & $\mathbf{0.730}_{.027}\,(3.0)$ & $\mathbf{0.852}_{.008}$ & $0.846_{.006}$ & $59.5_{1.1}$ & 538 \\
  \cmidrule(lr){2-12}
  & \multirow{3}{*}{100D}
    & \textsf{ATLAS-b} & $\mathbf{0.503}_{.052}$ & $\mathbf{62.5}_{11.3}$ & $45.2_{10.9}$ & & $0.586_{.150}\,(3.8)$ & $0.649_{.080}$ & $\mathbf{0.959}_{.027}$ & $22.3_{8.6}$ & 538 \\
  & & \textsf{ATLAS-s} & $0.015_{.011}$ & $0.8_{0.6}$ & $\mathbf{94.9}_{3.0}$ & & $0.657_{.030}\,(21.2)$ & $0.786_{.011}$ & $0.799_{.007}$ & $\mathbf{90.9}_{1.7}$ & 538 \\
  & & \textsf{ATLAS}   & $0.461_{.028}$ & $41.0_{5.7}$ & $55.0_{8.4}$ & & $\mathbf{0.844}_{.028}\,(32.8)$ & $\mathbf{0.914}_{.011}$ & $0.839_{.007}$ & $47.6_{2.8}$ & 538 \\
\bottomrule
\end{tabular}
\begin{tablenotes} 
\item[1] -- denotes precision undefined (no super-level predictions).
\end{tablenotes}
\end{threeparttable}
\end{table}

\begin{table}[t]
\centering
\caption{\textbf{ATLAS with default weights.} Same setup as Table~\ref{tab:joint-eval-multi-dim-main}, but with a single setting ($\boundaryweight{=}2$, $\superlevelprobweight{=}40$) for every benchmark. Subscripts are standard deviations. ATLAS still achieves the best recall in 8 of 9 experiments, the best F1 (Mix) in 7 of 9, and the highest depth (mV) in all 9, with dispersion within 30\% of the best.\tnote{1} Per-benchmark weights (Table~\ref{tab:joint-eval-multi-dim-main}) mainly improve Rastrigin 100D and the adversarial-set metrics.}
\label{tab:default-table}
\begin{threeparttable}
\scriptsize
\setlength{\tabcolsep}{1.5pt}
\begin{tabular}{lll ccc c ccccc}
\toprule
& & & \multicolumn{3}{c}{\textbf{Learned Model Metrics}} & & \multicolumn{5}{c}{\textbf{Sampling Set Metrics}} \\
\cmidrule(lr){4-6} \cmidrule(lr){8-12}
\textbf{Func} & \textbf{Dim} & \textbf{Method} & \textbf{F1 $\uparrow$} & \textbf{Rec $\superlevelset$ \% $\uparrow$} & \textbf{Prec\% $\uparrow$} & & \textbf{mV ($\%$) $\uparrow$} & \textbf{Margin J $\uparrow$} & \textbf{$\dispersionmetricnorm$ $\uparrow$} & \textbf{nSuper/B (\%) $\uparrow$} & \textbf{B} \\
\midrule

\multirow{9}{*}{\textbf{Ackley}} 
  & \multirow{3}{*}{10D}
    & AP2L   & $0.088_{.203}$ & $14.0_{29.9}$ & $49.3_{37.9}$ & & $0.096_{.033}\,(0.0)$ & $0.128_{.072}$ & $0.957_{.102}$ & $5.5_{9.1}$ & 650 \\
  & & TRLSE  & $\mathbf{0.681}_{.005}$ & $59.2_{1.5}$ & $\mathbf{80.2}_{1.5}$ & & $0.113_{.044}\,(0.0)$ & $0.042_{.011}$ & $\mathbf{1.019}_{.005}$ & $26.0_{2.0}$ & 650 \\
  & & ATLAS  & $0.664_{.020}$ & $\mathbf{63.6}_{4.9}$ & $69.8_{4.1}$ & & $\mathbf{0.304}_{.048}\,(0.0)$ & $\mathbf{0.368}_{.016}$ & $0.735_{.033}$ & $\mathbf{53.7}_{3.3}$ & 538 \\
  \cmidrule(lr){2-12}

  & \multirow{3}{*}{50D}
    & AP2L   & $0.010_{.024}$ & $0.5_{1.3}$ & $\mathbf{81.6}_{4.5}$ & & $0.021_{.009}\,(0.0)$ & $0.000_{.000}$ & $\mathbf{0.991}_{.009}$ & $2.9_{3.5}$ & 650 \\
  & & TRLSE  & $0.391_{.022}$ & $27.4_{2.4}$ & $68.6_{2.1}$ & & $0.035_{.008}\,(0.1)$ & $0.001_{.003}$ & $0.943_{.003}$ & $26.8_{1.2}$ & 650 \\
  & & ATLAS  & $\mathbf{0.454}_{.016}$ & $\mathbf{61.9}_{4.4}$ & $36.2_{3.2}$ & & $\mathbf{0.077}_{.009}\,(18.0)$ & $\mathbf{0.093}_{.018}$ & $0.784_{.007}$ & $\mathbf{52.0}_{1.1}$ & 538 \\
  \cmidrule(lr){2-12}

  & \multirow{3}{*}{100D}
    & AP2L   & $0.002_{.002}$ & $0.1_{0.1}$ & $7.0_{1.6}$ & & $0.010_{.003}\,(0.0)$ & $0.000_{.000}$ & $\mathbf{1.001}_{.004}$ & $3.5_{2.6}$ & 650 \\
  & & TRLSE  & $0.182_{.031}$ & $10.3_{2.0}$ & $\mathbf{79.0}_{2.4}$ & & $0.021_{.003}\,(0.1)$ & $0.000_{.000}$ & $0.928_{.002}$ & $39.2_{1.2}$ & 650 \\
  & & ATLAS  & $\mathbf{0.307}_{.038}$ & $\mathbf{20.8}_{3.6}$ & $60.6_{3.7}$ & & $\mathbf{0.051}_{.006}\,(20.3)$ & $\mathbf{0.006}_{.007}$ & $0.837_{.007}$ & $\mathbf{52.0}_{2.2}$ & 538 \\
\midrule

\multirow{9}{*}{\textbf{Rast.}} 
  & \multirow{3}{*}{10D}
    & AP2L   & $0.259_{.192}$ & $31.5_{30.5}$ & $53.3_{23.7}$ & & $0.474_{.110}\,(0.1)$ & $0.643_{.061}$ & $0.900_{.081}$ & $12.9_{11.4}$ & 650 \\
  & & TRLSE  & $0.366_{.034}$ & $26.0_{3.5}$ & $\mathbf{62.9}_{1.5}$ & & $0.489_{.051}\,(0.0)$ & $0.608_{.020}$ & $\mathbf{0.989}_{.012}$ & $30.0_{2.0}$ & 650 \\
  & & ATLAS  & $\mathbf{0.413}_{.043}$ & $\mathbf{34.5}_{7.2}$ & $54.2_{5.8}$ & & $\mathbf{0.600}_{.053}\,(0.0)$ & $\mathbf{0.720}_{.021}$ & $0.798_{.040}$ & $\mathbf{45.9}_{2.6}$ & 538 \\
  \cmidrule(lr){2-12}

  & \multirow{3}{*}{50D}
    & AP2L   & $0.245_{.201}$ & $21.1_{21.1}$ & $61.4_{14.0}$ & & $0.221_{.049}\,(0.0)$ & $0.388_{.067}$ & $\mathbf{0.991}_{.015}$ & $12.8_{11.7}$ & 650 \\
  & & TRLSE  & $0.396_{.019}$ & $27.1_{1.8}$ & $\mathbf{73.7}_{1.3}$ & & $0.257_{.032}\,(0.0)$ & $0.443_{.015}$ & $0.952_{.004}$ & $35.2_{1.5}$ & 650 \\
  & & ATLAS  & $\mathbf{0.470}_{.024}$ & $\mathbf{38.8}_{4.6}$ & $60.9_{3.3}$ & & $\mathbf{0.339}_{.029}\,(0.9)$ & $\mathbf{0.541}_{.021}$ & $0.989_{.021}$ & $\mathbf{42.2}_{2.6}$ & 538 \\
  \cmidrule(lr){2-12}

  & \multirow{3}{*}{100D}
    & AP2L   & $\mathbf{0.227}_{.144}$ & $\mathbf{14.9}_{9.6}$ & $67.1_{2.7}$ & & $0.150_{.029}\,(0.0)$ & $0.276_{.054}$ & $0.995_{.005}$ & $15.5_{10.2}$ & 650 \\
  & & TRLSE  & $0.000_{.000}$ & $0.0_{0.0}$ & -- & & $0.183_{.021}\,(0.0)$ & $0.327_{.015}$ & $0.922_{.002}$ & $\mathbf{47.1}_{3.1}$ & 650 \\
  & & ATLAS  & $0.066_{.030}$ & $3.5_{1.7}$ & $\mathbf{78.6}_{6.5}$ & & $\mathbf{0.203}_{.037}\,(0.1)$ & $\mathbf{0.367}_{.046}$ & $\mathbf{1.089}_{.037}$ & $29.0_{8.7}$ & 538 \\
\midrule

\multirow{9}{*}{\textbf{Rosen.}} 
  & \multirow{3}{*}{10D}
    & AP2L   & $0.019_{.044}$ & $10.5_{30.4}$ & $50.3_{49.7}$ & & $0.676_{.127}\,(0.0)$ & $0.742_{.168}$ & $0.993_{.030}$ & $4.0_{3.1}$ & 650 \\
  & & TRLSE  & $0.682_{.016}$ & $57.6_{2.3}$ & $\mathbf{83.8}_{2.0}$ & & $0.812_{.059}\,(0.0)$ & $0.811_{.015}$ & $\mathbf{1.086}_{.015}$ & $\mathbf{32.8}_{5.8}$ & 650 \\
  & & ATLAS  & $\mathbf{0.693}_{.013}$ & $\mathbf{79.3}_{3.9}$ & $61.9_{4.0}$ & & $\mathbf{0.877}_{.051}\,(0.0)$ & $\mathbf{0.910}_{.020}$ & $0.952_{.028}$ & $28.8_{2.8}$ & 538 \\
  \cmidrule(lr){2-12}

  & \multirow{3}{*}{50D}
    & AP2L   & $0.006_{.024}$ & $3.6_{18.3}$ & $60.7_{45.8}$ & & $0.326_{.058}\,(0.0)$ & $0.528_{.039}$ & $\mathbf{0.997}_{.007}$ & $3.8_{2.8}$ & 650 \\
  & & TRLSE  & $0.185_{.024}$ & $10.6_{1.5}$ & $\mathbf{75.2}_{2.2}$ & & $0.536_{.053}\,(0.0)$ & $\mathbf{0.712}_{.015}$ & $0.921_{.004}$ & $\mathbf{37.2}_{2.2}$ & 650 \\
  & & ATLAS  & $\mathbf{0.498}_{.024}$ & $\mathbf{43.8}_{4.7}$ & $58.5_{3.5}$ & & $\mathbf{0.567}_{.071}\,(0.1)$ & $0.680_{.026}$ & $0.955_{.008}$ & $17.3_{1.3}$ & 538 \\
  \cmidrule(lr){2-12}

  & \multirow{3}{*}{100D}
    & AP2L   & $0.006_{.022}$ & $3.6_{18.3}$ & $4.1_{3.8}$ & & $0.250_{.055}\,(0.0)$ & $0.417_{.065}$ & $\mathbf{0.998}_{.008}$ & $3.5_{2.6}$ & 650 \\
  & & TRLSE  & $0.000_{.000}$ & $0.0_{0.0}$ & -- & & $0.397_{.025}\,(0.0)$ & $0.603_{.011}$ & $0.937_{.001}$ & $\mathbf{66.3}_{1.7}$ & 650 \\
  & & ATLAS  & $\mathbf{0.371}_{.020}$ & $\mathbf{67.4}_{4.8}$ & $\mathbf{25.7}_{2.1}$ & & $\mathbf{0.672}_{.153}\,(21.0)$ & $\mathbf{0.797}_{.122}$ & $0.834_{.057}$ & $13.4_{5.0}$ & 538 \\
\bottomrule
\end{tabular}
\begin{tablenotes} 
\item[1] Sampling deep inside the super level set sacrifices average dispersion, since samples there are closer together than samples along the boundary. As ATLAS goes deep into the super level set to produce a set with more representative margin spread, its set’s coverage decreases on average, which is why our dispersion spread is somewhat lower than our baselines. Although it loses coverage, it gains coverage of adversarial scores that would not otherwise be recoverable with the other baselines or uniform sampling. At 50D and 100D, ATLAS recovers adversarial scores that are higher than every sample in the super level test set.
\end{tablenotes}
\end{threeparttable}
\end{table}

\section{Functions Used In Toy Experiments}
\label{Appendix: Function}
\begin{figure*}[t] 
    \centering
    \begin{subfigure}[b]{0.30\textwidth}
      \centering
      \includegraphics[width=\textwidth]{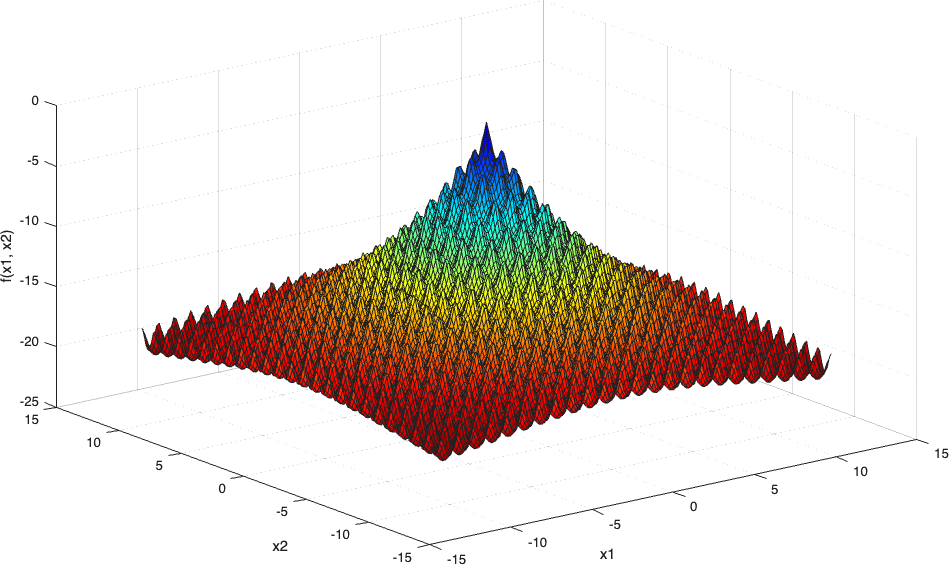}
      \caption{Ackley Function}
      \label{fig: ackley}
    \end{subfigure}
    \hfill 
    \begin{subfigure}[b]{0.30\textwidth}
      \centering
      \includegraphics[width=\textwidth]{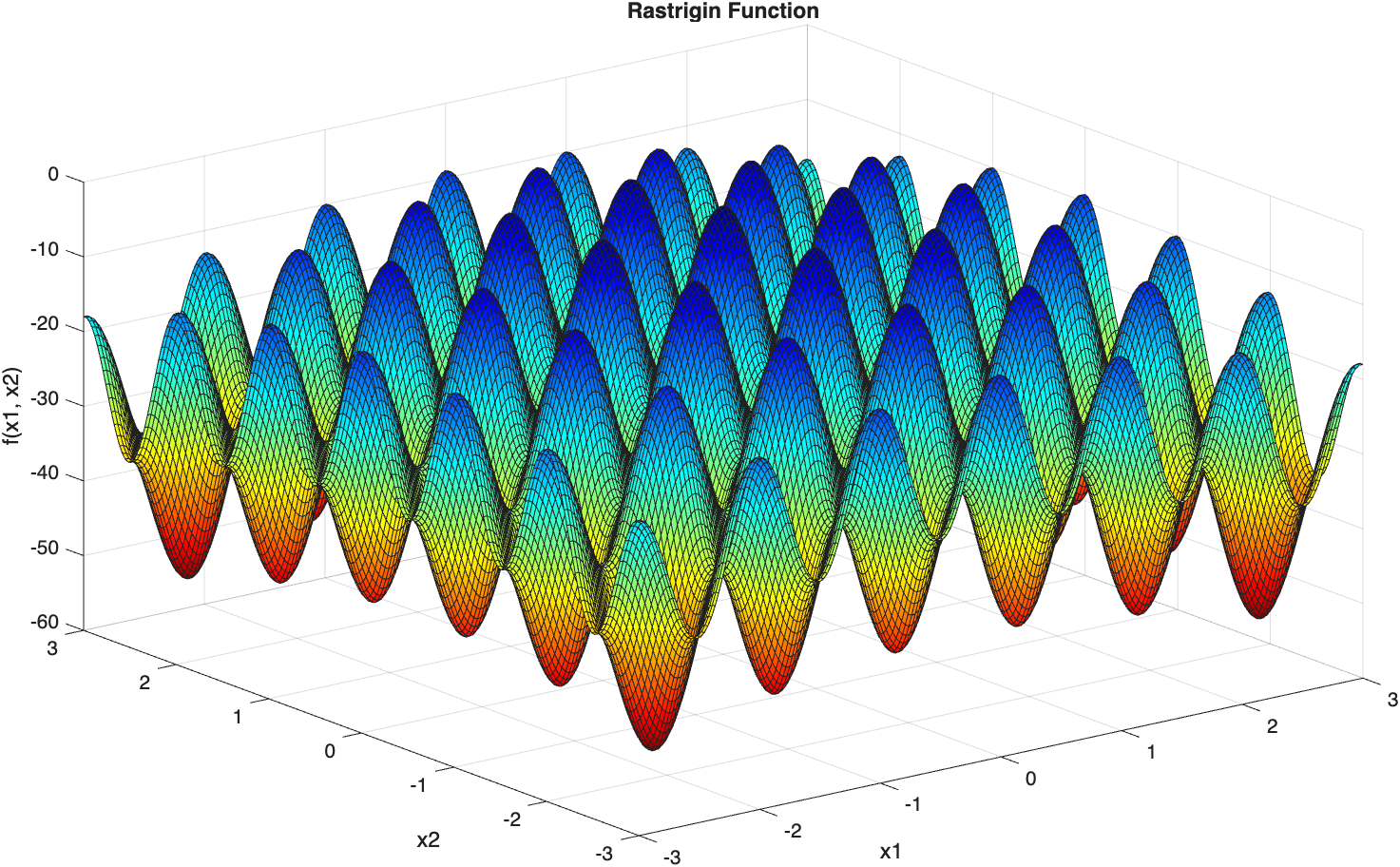}
      \caption{Rastrigin Function}
      \label{fig: rastrigin}
    \end{subfigure}
    \hfill
    \begin{subfigure}[b]{0.30\textwidth}
      \centering
      \includegraphics[width=\textwidth]{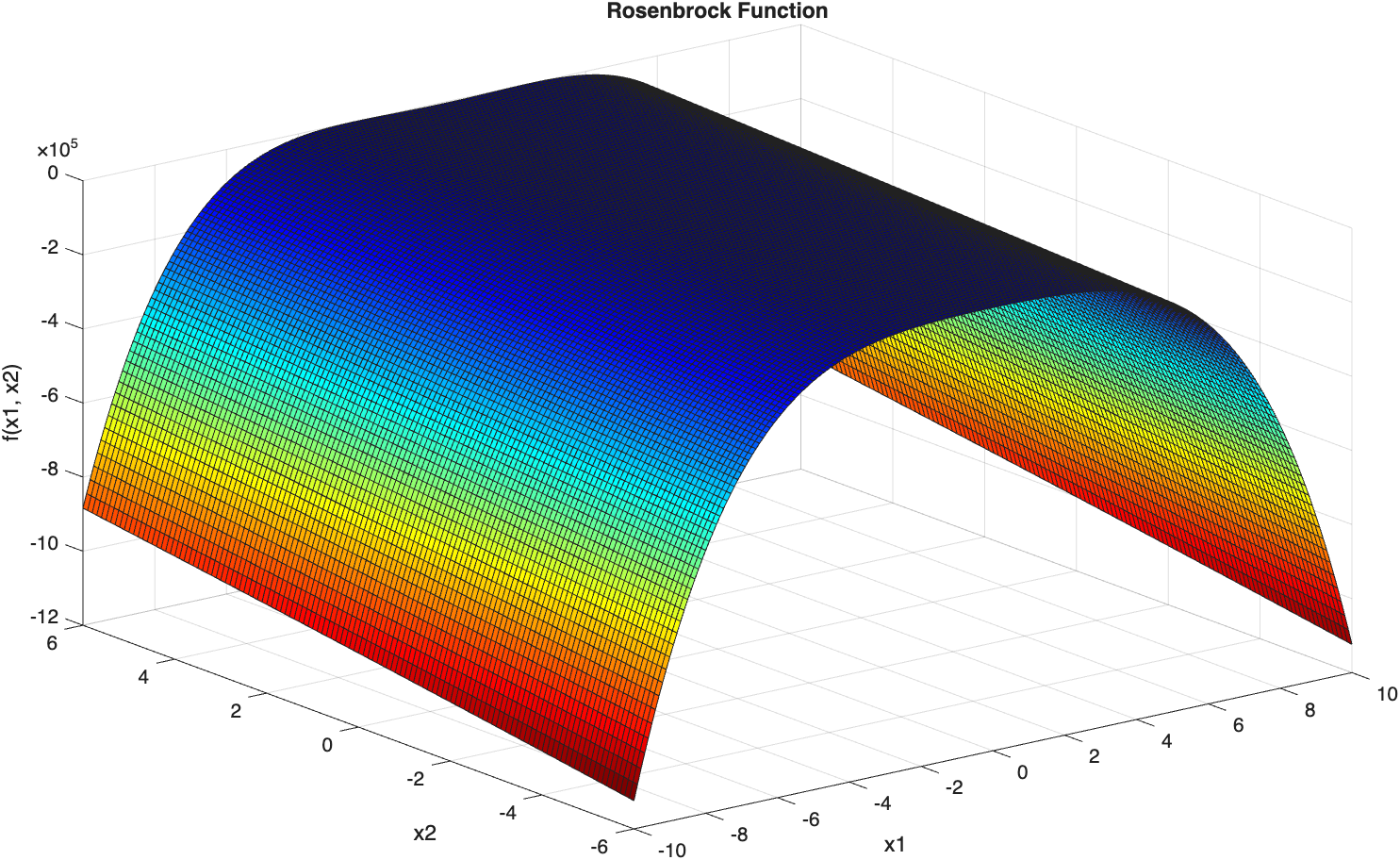}
      \caption{Rosenbrock Function}
      \label{fig: rosenbrock}
    \end{subfigure}
    \hfill
    \caption{Non-convex toy functions used in our toy experiments described in Section \ref{sec: toy_exp}.}
    \label{fig:toyfunctions}
\end{figure*}

\subsection{Ackley Function}
This function is described by the following:
\begin{linenomath}
\begin{align}
\label{eq: ackley}
    \toyfunction(\mathbf{\toyinput}) = &-\ackleyconstanta \exp \left( -\ackleyconstantb \sqrt{\frac{1}{\toyinputdim} \sum_{\toyinputidx=1}^{\toyinputdim} \toyinput_\toyinputidx^2}\right) \\
    &- \exp \left( \frac{1}{\toyinputdim} \sum_{\toyinputidx=1}^{\toyinputdim} \cos(\ackleyconstantc \toyinput_\toyinputidx) \right) + \ackleyconstanta + \exp(1) \notag
\end{align}
\end{linenomath}
where $\mathbf{\toyinput} \in \mathbf{R}^{\toyinputdim}$, and $\ackleyconstanta=20$, $\ackleyconstantb = 0.2$, and $\ackleyconstantc = 2\pi$. The global minimum has value 0 at $\mathbf{\toyinput} = \mathbf{0}$.

\subsection{Rastrigin Function}
This function is described by the following:
\begin{linenomath}
\begin{align}
\label{eq: rastrigin}
    \toyfunction(\mathbf{\toyinput}) = \rastriginconstanta \toyinputdim + \sum_{\toyinputidx=1}^{\toyinputdim} \left[ \toyinput_\toyinputidx^2 - \rastriginconstanta \cos(2\pi \toyinput_\toyinputidx) \right]
\end{align}
\end{linenomath}
where $\mathbf{\toyinput} \in \mathbf{R}^{\toyinputdim}$, and $\rastriginconstanta = 10$. The global minimum has value 0 at $\mathbf{\toyinput} = \mathbf{0}$.

\subsection{Rosenbrock Function}
This function is described by the following:
\begin{linenomath}
\begin{align}
\label{eq: rosenbrock}
    \toyfunction(\mathbf{\toyinput}) = \sum_{\toyinputidx=1}^{\toyinputdim-1} \left[ \rosenbrockconstantb (\toyinput_{\toyinputidx+1} - \toyinput_\toyinputidx^2)^2 + (\rosenbrockconstanta - \toyinput_\toyinputidx)^2 \right]
\end{align}
\end{linenomath}
where $\mathbf{\toyinput} \in \mathbf{R}^{\toyinputdim}$, and $\rosenbrockconstanta = 1$, $\rosenbrockconstantb = 100$. The global minimum has value 0 at $\mathbf{\toyinput} = \mathbf{1} = (1, \dots, 1)$.

\section{Implementation Speed up}
\label{sec: Speedups}
In large dimensions, most of ATLAS’s runtime comes from training a new global model on samples and choosing the new TR centers using limited memory BFGS with bound constraints (L-BFGS-B) and warm starts. To address the latter, we ultimately use analytic gradients during optimization\footnote{We experiment between using analytic gradient of our placement acquisition functions and using batched finite difference as gradients. While each have their specific speed up benefits, the former displayed higher speeds for larger dimensions during longer runs.} and parallelize specific calculations. Both mechanisms give us tremendous speed ups, which are most beneficial after the model starts acquiring a large amount of samples during its run. The timing metrics shown are runtimes when using CPU. Depending on the choice of model and learning algorithm, further speeds up can be realized by using GPU resources.

\section{Image Experiment Details}
\label{sec:Image Appendix}
The models chosen are ResNet50 \citep{he2016deep} for CIFAR's and ImageNet's standard target with 93.65\% and 76.13\% accuracy respectively, WRN-28-10 adversarially trained (\textbf{AT}) with DDPM \citep{rebuffi2021fixing} with 60.75\% accuracy for CIFAR's \textbf{AT} target, ResNet50 \textbf{AT} on $L_{inf}$ perturbations \citep{salman2020adversarially} with 54.53\% accuracy for ImageNet's \textbf{AT} target, and a pre-trained LeNet5 from \citep{singh2019gan} with 99\% accuracy. We use a conditional DCGAN \citep{radford2015unsupervised} and BIGGAN and BIGGAN-256 \citep{brock2018large} for our image generators for MNIST, CIFAR, and ImageNet respectively. BIGGAN and the conditional DCGAN was retrained from scratch using \citep{singh2019ganvaepretrained} implementation to take in 32 dimensions and a source class, while the implementation and trained weights for BIGGAN-256 comes from StudioGAN\citep{kang2023studiogan}. BayesOpt is run in its native early-stopping configuration, while NES and SignHunter is ran until the full budget generating its version of an adversarial set. We tried to run BayesOpt similarly; however, we found it to be more computationally infeasible to do. We therefore report ASR and query-to-success for BayesOpt, and omit the dispersion statistics, which require the run-to-budget super-level sets. Full analysis of ATLAS, NES, SignHunter, BayesOpt, and uniform sampling is in Table \ref{tab:image-latent-main-appendix}. 

\begin{table}[t]
\centering
\caption{\textbf{At a matched query budget, ATLAS returns better distributed adversarial sets than the query-based attacks, at comparable attack success.} We compare ATLAS to NES, SignHunter, BayesOpt, and uniform random sampling at finding generated images under a source class that gets misclassified into a target class by the target model. We average each metric over 20 different source-target pairs per target model, and give each method a budget of $B=650$ oracle queries per experiment. Subscripts are standard deviations across pairs. Best per metric and target in bold. ASR is the fraction of pairs for which the method finds at least one adversarial example. $Q_{first}$ is the median queries to the first adversarial example and is computed over the pairs on which the method succeeded. Thus, it is not comparable across methods with different ASR. `Adv. Count' is the number of adversarial examples found. Margin J, $\dispersionmetricnorm$ and $\clump$ describe how those points are distributed (higher = more of the region covered, and covered more evenly). At 650-budget, ATLAS recovers the most representative adversarial set. It is best on Margin J in 4 of 5 targets with good spatial spread and spacing. ATLAS recovers fewer raw points than the query-based attacks but distributes them far better. The other query baselines are good at finding adversarial examples but poor at characterizing the adversarial region. SignHunter finds the most points, yet its $\clump$ of 0.24--0.31 versus ATLAS's 0.55--0.65 shows they are in tight clusters; NES's $\dispersionmetricnorm$ of 1.14--2.43 versus ATLAS's 1.51--3.82 shows its points are in smaller neighbourhoods than ATLAS. Uniform random sampling has more spread by construction, but finds only 1--19 points on average.}
\label{tab:image-latent-main-appendix}
\begin{threeparttable}
\scriptsize
\setlength{\tabcolsep}{1.5pt}
\begin{tabular}{ll l cc c cccc}
\toprule
& & & \multicolumn{2}{c}{\textbf{Attack}} & & \multicolumn{4}{c}{\textbf{Adversarial Set}} \\
\cmidrule(lr){4-5} \cmidrule(lr){7-10}
\textbf{Data} & \textbf{Target} & \textbf{Method} & \textbf{ASR\% $\uparrow$} & \textbf{$Q_{first}$ $\downarrow$} & & \textbf{Margin J $\uparrow$} & \textbf{$\dispersionmetricnorm$ $\uparrow$} & \textbf{Adv. Count $\uparrow$} & \textbf{$\clump$ $\uparrow$} \\
\midrule

  \multirow{5}{*}{\textbf{MNIST}} & \multirow{5}{*}{Standard} & \textbf{ATLAS}         & $95$ & $164_{85}$ & & $\mathbf{0.859}_{0.091}$ & $1.94_{0.15}$ & $174.4_{109.1}$ & $0.63_{0.04}$ \\
   &  & NES                    & $75$ & $181_{162}$ & & $0.745_{0.238}$ & $1.14_{0.12}$ & $225.9_{199.2}$ & $0.58_{0.13}$ \\
   &  & SignHunter             & $\mathbf{100}$ & $60_{48}$ & & $0.816_{0.075}$ & $\mathbf{2.99}_{0.31}$ & $\mathbf{255.4}_{99.4}$ & $0.27_{0.03}$ \\
   &  & Uniform                & $26$ & $408_{211}$ & & $0.136_{0.207}$ & $2.25_{0.03}$ & $1.2_{3.5}$ & $\mathbf{0.92}_{0.10}$ \\
   &  & BayesOpt\tnote{a}      & $\mathbf{100}$ & $\mathbf{36}_{90}$ & & -- & -- & -- & -- \\
  \midrule
  \multirow{5}{*}{\textbf{CIFAR-10}} & \multirow{5}{*}{Standard} & \textbf{ATLAS}         & $95$ & $203_{139}$ & & $\mathbf{0.818}_{0.200}$ & $1.51_{0.30}$ & $71.8_{64.7}$ & $0.63_{0.13}$ \\
   &  & NES                    & $65$ & $143_{214}$ & & $0.804_{0.267}$ & $1.18_{0.15}$ & $94.2_{119.4}$ & $0.57_{0.10}$ \\
   &  & SignHunter             & $\mathbf{100}$ & $72_{104}$ & & $0.762_{0.078}$ & $\mathbf{2.92}_{0.41}$ & $\mathbf{256.6}_{122.8}$ & $0.27_{0.09}$ \\
   &  & Uniform                & $60$ & $106_{103}$ & & $0.492_{0.259}$ & $2.23_{0.15}$ & $5.2_{9.0}$ & $\mathbf{0.88}_{0.07}$ \\
   &  & BayesOpt\tnote{a}      & $95$ & $\mathbf{57}_{54}$ & & -- & -- & -- & -- \\
  \midrule
  \multirow{5}{*}{\textbf{CIFAR-10}} & \multirow{5}{*}{Adv.} & \textbf{ATLAS}         & $\mathbf{100}$ & $83_{81}$ & & $\mathbf{0.823}_{0.114}$ & $1.88_{0.18}$ & $203.4_{96.9}$ & $0.57_{0.03}$ \\
   &  & NES                    & $95$ & $121_{95}$ & & $0.803_{0.162}$ & $1.24_{0.13}$ & $282.6_{172.2}$ & $0.52_{0.08}$ \\
   &  & SignHunter             & $\mathbf{100}$ & $47_{67}$ & & $0.770_{0.109}$ & $\mathbf{3.00}_{0.31}$ & $\mathbf{306.6}_{142.4}$ & $0.24_{0.05}$ \\
   &  & Uniform                & $85$ & $96_{163}$ & & $0.356_{0.204}$ & $2.24_{0.09}$ & $10.9_{15.8}$ & $\mathbf{0.87}_{0.07}$ \\
   &  & BayesOpt\tnote{a}      & $\mathbf{100}$ & $\mathbf{44}_{45}$ & & -- & -- & -- & -- \\
  \midrule
  \multirow{5}{*}{\textbf{ImageNet}} & \multirow{5}{*}{Standard} & \textbf{ATLAS}         & $85$ & $229_{168}$ & & $0.609_{0.272}$ & $3.33_{0.93}$ & $86.4_{112.8}$ & $0.65_{0.16}$ \\
   &  & NES                    & $90$ & $\mathbf{97}_{200}$ & & $\mathbf{0.720}_{0.250}$ & $2.41_{0.34}$ & $175.6_{207.4}$ & $0.63_{0.13}$ \\
   &  & SignHunter             & $\mathbf{95}$ & $151_{176}$ & & $0.538_{0.242}$ & $2.47_{1.16}$ & $\mathbf{184.1}_{94.8}$ & $0.31_{0.17}$ \\
   &  & Uniform                & $60$ & $107_{185}$ & & $0.342_{0.262}$ & $\mathbf{4.49}_{0.13}$ & $18.9_{40.7}$ & $\mathbf{0.93}_{0.06}$ \\
   &  & BayesOpt\tnote{a}      & $90$\tnote{b} & $130_{121}$ & & -- & -- & -- & -- \\
  \midrule
  \multirow{5}{*}{\textbf{ImageNet}} & \multirow{5}{*}{Adv.} & \textbf{ATLAS}         & $\mathbf{100}$ & $162_{158}$ & & $\mathbf{0.745}_{0.206}$ & $3.82_{0.75}$ & $115.1_{104.4}$ & $0.55_{0.14}$ \\
   &  & NES                    & $85$ & $98_{133}$ & & $0.703_{0.194}$ & $2.43_{0.23}$ & $144.6_{180.2}$ & $0.61_{0.09}$ \\
   &  & SignHunter             & $85$ & $147_{150}$ & & $0.583_{0.189}$ & $3.37_{0.97}$ & $\mathbf{195.4}_{112.1}$ & $0.26_{0.08}$ \\
   &  & Uniform                & $55$ & $\mathbf{25}_{118}$ & & $0.391_{0.264}$ & $\mathbf{4.48}_{0.12}$ & $15.3_{39.9}$ & $\mathbf{0.93}_{0.05}$ \\
   &  & BayesOpt\tnote{a}      & $\mathbf{100}$ & $102_{121}$ & & -- & -- & -- & -- \\

\bottomrule
\end{tabular}
\begin{tablenotes}
\item[a] BayesOpt is run in its published configuration, which stops at its first
adversarial point, so it is read for ASR and $Q_{first}$ only: its Count is $1$ by
construction. 
\item[b] 2 of the 20 batches for BayesOpt targeting the standard ImageNet were terminated. We cap a
single batch at ${\sim}12$\,h of wall-clock time, which they exceeded having completed only
${\sim}490$ and ${\sim}585$ of 650 queries. Neither had found an adversarial point, and both are counted as failures.
No other batch of any arm reached this cap. The slowest ATLAS batch took 37\,min, and no NES or SignHunter batch exceeded 2\,min. 
This illustrates the computational inefficiency of BayesOpt in this setting. 
\end{tablenotes}
\end{threeparttable}
\end{table}

\section{Interpreting ATLAS Image Results}
\label{sec:interpret_image_results}
\begin{figure*}[t]
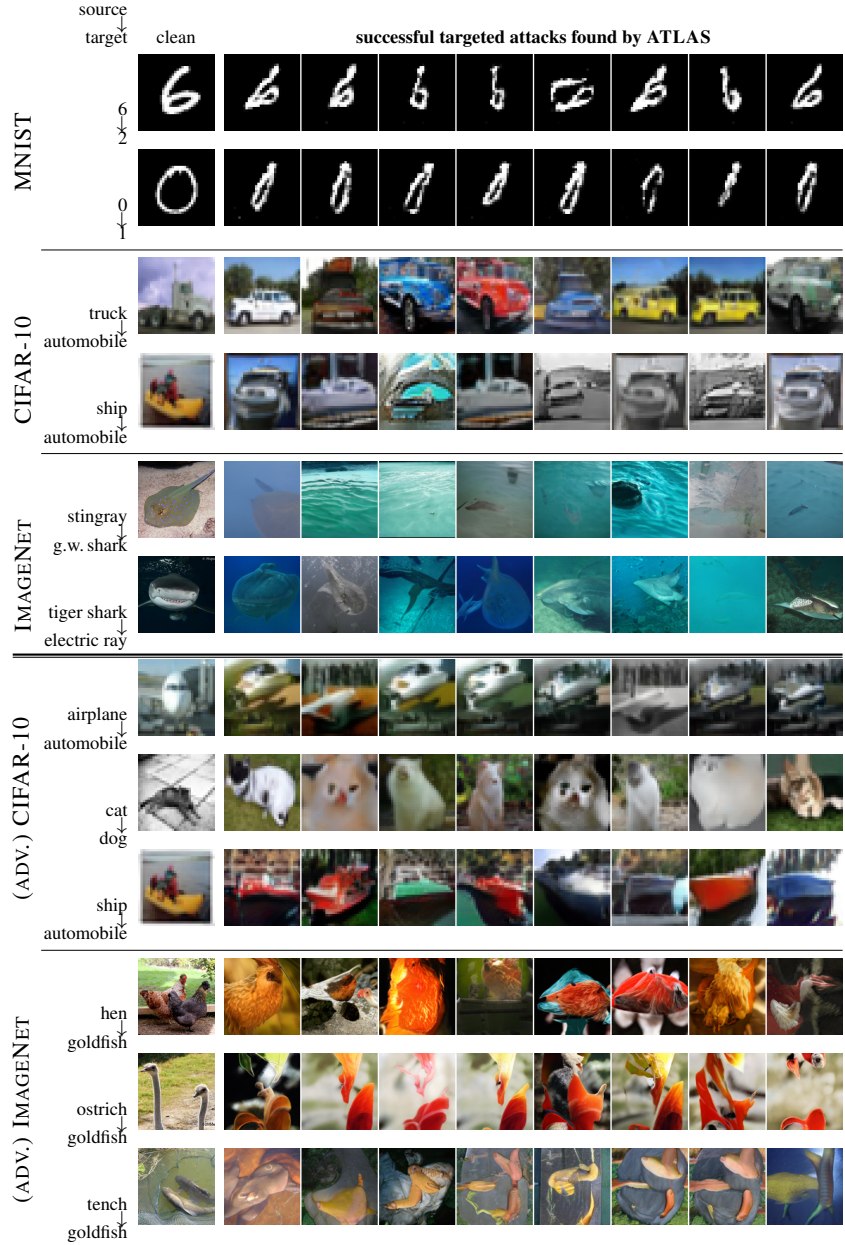

  \centering
  {
  \setlength{\tabcolsep}{2.0pt}
  \renewcommand{\arraystretch}{0.92}

  \begin{tabular}{@{}c r c c@{}}

    &
    \scriptsize
    \begin{tabular}[b]{@{}r@{}}
      source\\[-2pt]
      $\downarrow$\\[-2pt]
      target
    \end{tabular}
    &
    \scriptsize clean
    &
    \scriptsize\textbf{successful targeted attacks found by \textsc{ATLAS}}
    \tabularnewline[1pt]


    \multirow{2}{*}{\dslabel{MNIST}}
    & \pairlabel{6}{2}
    & \atlasclean{standard_mnist_b4}
    & \atlasadv{standard_mnist_b4}
    \tabularnewline


    & \pairlabel{0}{1}
    & \atlasclean{standard_mnist_b1}
    & \atlasadv{standard_mnist_b1}
    \tabularnewline

    \cmidrule(l{1pt}r{0pt}){2-4}

    \multirow{2}{*}{\dslabel{CIFAR-10}}
    & \pairlabel{truck}{automobile}
    & \atlasclean{standard_cifar10_b0}
    & \atlasadv{standard_cifar10_b0}
    \tabularnewline

    & \pairlabel{ship}{automobile}
    & \atlasclean{standard_cifar10_b2}
    & \atlasadv{standard_cifar10_b2}
    \tabularnewline


    \cmidrule(l{1pt}r{0pt}){2-4}

    \multirow{2}{*}{\dslabel{ImageNet}}
    & \pairlabel{stingray}{g.w.\,shark}
    & \atlasclean{standard_imagenet_b4}
    & \atlasadv{standard_imagenet_b4}
    \tabularnewline

    & \pairlabel{tiger shark}{electric ray}
    & \atlasclean{standard_imagenet_b3}
    & \atlasadv{standard_imagenet_b3}
    \tabularnewline

    \specialrule{0.8pt}{0.5pt}{0.2pt}
    \specialrule{0.4pt}{0.2pt}{0.5pt}

    \multirow{3}{*}{\dslabel{(adv.) CIFAR-10}}
    & \pairlabel{airplane}{automobile}
    & \atlasclean{adv_trained_cifar10_b1}
    & \atlasadv{adv_trained_cifar10_b1}
    \tabularnewline

    & \pairlabel{cat}{dog}
    & \atlasclean{adv_trained_cifar10_b3}
    & \atlasadv{adv_trained_cifar10_b3}
    \tabularnewline

    & \pairlabel{ship}{automobile}
    & \atlasclean{adv_trained_cifar10_b2}
    & \atlasadv{adv_trained_cifar10_b2}
    \tabularnewline

    \cmidrule(l{1pt}r{0pt}){2-4}

    \multirow{3}{*}{\dslabel{(adv.) ImageNet}}
    & \pairlabel{hen}{goldfish}
    & \atlasclean{adv_trained_imagenet_b2}
    & \atlasadv{adv_trained_imagenet_b2}
    \tabularnewline

    & \pairlabel{ostrich}{goldfish}
    & \atlasclean{adv_trained_imagenet_b0}
    & \atlasadv{adv_trained_imagenet_b0}
    \tabularnewline

    & \pairlabel{tench}{goldfish}
    & \atlasclean{adv_trained_imagenet_b1}
    & \atlasadv{adv_trained_imagenet_b1}
    \tabularnewline

  \end{tabular}
  }

  \vspace{1pt}

  \caption{
    Representative adversarial examples discovered by \textsc{ATLAS} under a 650-query budget. Each row specifies a source and target class. The leftmost image is a clean source-class sample, followed by representative samples from the generated adversarial set. The examples for MNIST, CIFAR, and (adv.) CIFAR contains adversarial images which could present as benign but that gets misclassified by the model. The examples for ImageNet and (adv.) ImageNet ALL contain distinct visual cues which pushes that model’s decision boundary towards that specific class. For the great white (g.w.) shark, the model is most likely to key in on the ocean (a shark’s environment) rather than a shark’s features for its classification. For goldfish, this model accepts non-goldfish structures as reasonable substitutes (a flower petal) when enough lower-level or mid-level goldfish visual cues are present. See Section \ref{sec:interpret_image_results} for details. 
  }

  \label{fig:atlas-examples}
\end{figure*}

For each experiment, we ask: (1) whether the generated adversarial sets contain images with visual semantics that are common under the source class with little to no features of the target class or (2) whether there are any particular characteristics or patterns that emerge from all images (realistic or not) that could indicate which features the model's boundary region for the target class is built on. The generated adversarial sets for both MNIST and CIFAR represent the former. These are interesting failure cases because it represents manifold-constrained adversarial images which could present as benign or containing imperceptible perturbations but that gets misclassified by the model (most similar to $L_{\infty}$ attacks). On the other hand, the generated adversarial sets for ImageNet represent more of the latter. The latter is important analysis because it tells us what features the models may hone in on for a particular classification and tells us what features push the model’s decision boundary from one class to another. Take as example the “stingray → great white shark (g.w. shark)” row in the ``Imagenet'' section in Figure \ref{fig:atlas-examples}. All images contain a birds-eye view of the ocean with sightings of a stingray in the water which are all reasonable images for the stingray class. Because these images are being classified as a great white shark, this tells us that the model is most likely to key in on the ocean (a shark’s environment) rather than a shark’s features for its classification. As another example, take every row in the ``(Adv.) ImageNet'' section displaying adversarial sets for a targeted goldfish classification. The images that ATLAS recovers contains both visual features that you’d expect to see in a goldfish image - orange/gold coloration, strong high-contrast colors, curved/elongated flowing structures, structures that vaguely resemble fins - and visual features that you would not expect in a typical goldfish image - structures that resembles flowers or petals, feathers, bird parts, etc. This suggests that the model may accept non-goldfish structures as reasonable substitutes when enough lower-level or mid-level goldfish visual cues are present. For example, \textit{an image with a flower petal or feather may be accepted as an image substitute because it provides a structure that is visually similar to a fin without actually being a fin}. This sort of observation can be used as a guideline for producing more sets of adversarial image failures for this particular model. 

\section{Security Audit Summary}
\label{sec:security_audit}

We interpret \textsc{ATLAS} as a model-level security auditing tool and summarize our audit findings using the NIST AI Risk Management Framework (AI RMF) \citep{ai2023artificial} and the MITRE ATLAS threat framework \citep{mitreatlas}. The evaluation models an attacker who can query a classifier but does not require access to its weights or training infrastructure. In MITRE ATLAS terminology, the procedure generates \emph{adversarial data} (AML.T0043) to \emph{evade an AI model} (AML.T0015). The potential consequences of successful misclassification are context dependent and should be assessed separately. 

\textbf{MAP.} We identify targeted adversarial attacks that represent security risks using a chosen image generator. We characterize ``failure regions’’ within this generator’s input latent space that contain latents that map to images containing the visual features of a specified source class but for which gets misclassified by the target model. 

\textbf{MEASURE.} \textsc{ATLAS} measures the accessibility and size of each discovered failure region under a fixed query budget (see Table \ref{tab:atlas_audit_summary}). Across all reported settings, \textsc{ATLAS} recovered nonempty sets, including against adversarially trained CIFAR-10 and ImageNet models, indicating that the adversarial training process for each adversarially trained model did not eliminate the existence of targeted failure regions. 

\textbf{MANAGE.} The resulting adversarial sets point to model vulnerabilities that are fairly accessible for outside users. The examples within these sets should be used to guide additional robustness evaluation, adversarial training, or the creation of monitoring and security filters as potential defenses.

\textbf{GOVERN.} Finally, the recovered adversarial sets, ATLAS’s trained surrogate model, and quantitative measurements, all encodes the state of security for each subject target, while the threat model, query budget, source-target pairs, and experimental settings provide documentation for how to repeat this security analysis after the model has been robustified with better security defenses. Each of these resulting audit artifacts should be used to inform deployment decisions.  
\section*{Ethical Considerations}
This system, while designed for developers to robustify their own systems, may be leveraged by outside actors to attack other systems they do not own. While this may be the case, to mitigate the potential damage of such cases, future work includes analyzing the limitations of this system, which includes finding the functional characteristics (defenses) for which similar algorithms are unable to efficiently map inputs their associated adversarial scores. Other possible mitigations include limiting query access to outside systems. 

\section*{AI Statement}
AI was used to help run smaller debugging experiments, analysis, and sweeps for parameter settings; help set up experimental scripts; help create figures and assist with writing revising; and give feedback on the authors methodological ideas and data interpretations. All core experimental, methodological, and analytical designs were done by the authors, along with their scripts; however, bug editing and enhancements for parallelization was assisted with AI. Core analysis was done by the authors from logs and data generated by the authors with assistance from AI. All code and any artifacts generated by AI were looked over and edited by the authors.

\section*{Open Science}
Complete hyperparameters for ATLAS and all baselines, including shared, matched, and method-specific settings, are given in Table \ref{tab:hyperparams}; experimental protocols are in Appendices \ref{sec: appendix-ext} and \ref{sec:Image Appendix}. All source code, configurations, and scripts to reproduce every reported result will be released publicly upon publication.

\end{document}